\documentclass[11pt]{article}

\usepackage[final]{acl}
\usepackage{natbib}

\usepackage{times}
\usepackage{latexsym}

\usepackage[T1]{fontenc}

\usepackage[utf8]{inputenc}

\usepackage{microtype}

\usepackage{inconsolata}

\usepackage{graphicx}
\usepackage{url}            
\usepackage{booktabs}       
\usepackage{amsfonts}       
\usepackage{nicefrac}       
\usepackage{microtype}      
\usepackage{xcolor}
\usepackage{xspace}

\usepackage{enumitem}
\usepackage{fontawesome5}
\usepackage{simpleicons}
\usepackage{graphicx}
\usepackage{xspace}
\usepackage{amssymb}
\usepackage{amsmath}
\usepackage{algorithm}
\usepackage{algpseudocode}
\usepackage{adjustbox}
\usepackage{verbatim}
\usepackage{fvextra}
\usepackage{listings}
\usepackage{fancyvrb}
\usepackage{cleveref}
\usepackage{titlesec}
\usepackage{caption}
\usepackage{multicol}
\usepackage{tablefootnote}
\titlespacing*{\paragraph}{0pt}{0pt}{1em}
\usepackage{booktabs,tabularx,array,pifont}
\newcommand{\cmark}{\ding{51}}  
\newcommand{\xmark}{\ding{55}}  
\newcolumntype{Y}{>{\raggedright\arraybackslash}X} 
\newenvironment{appendixtext}{\begin{multicols}{2}\raggedcolumns}{\end{multicols}}

\newcommand{\huggingfacelogo}{\raisebox{-0.2ex}{\includegraphics[height=1.2em]{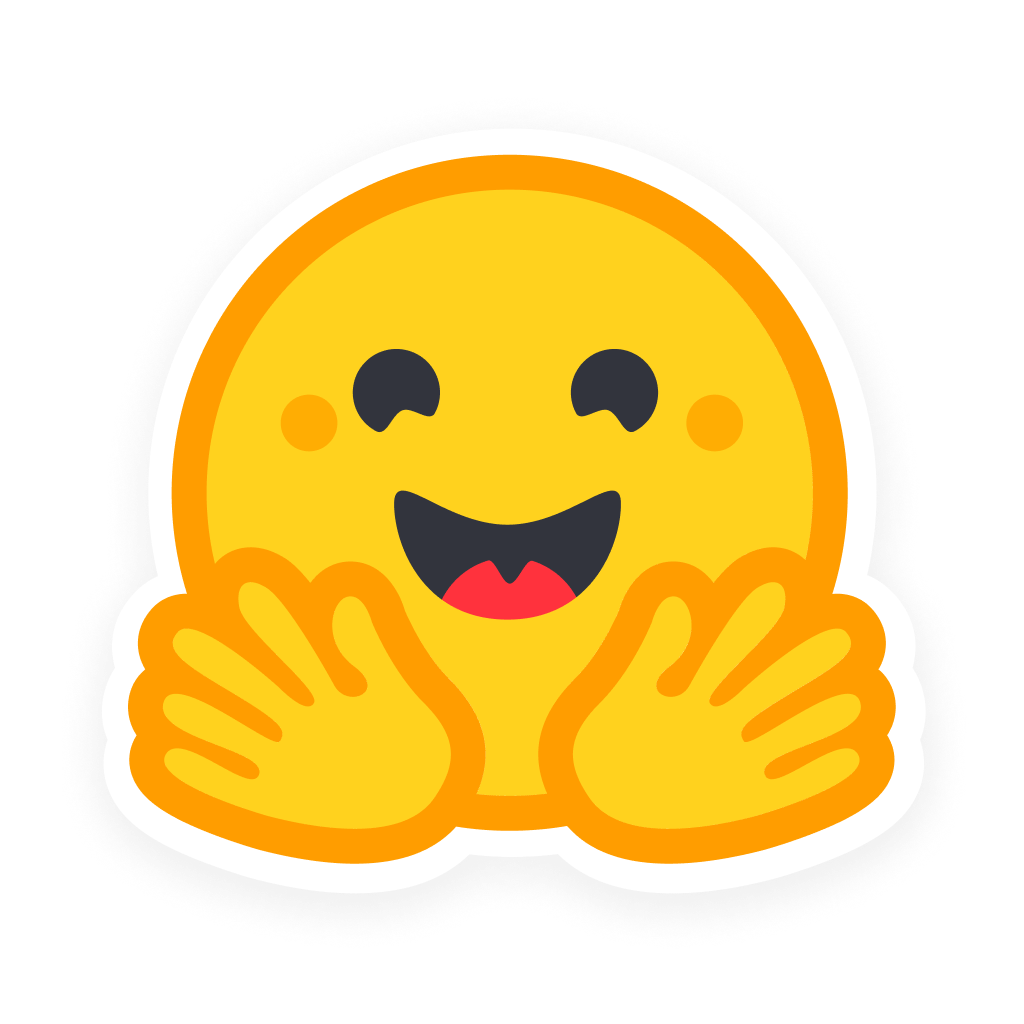}}\xspace}

\crefname{section}{Sec.}{Sec.}
\crefname{theorem}{Theorem}{Theorems}
\crefname{corollary}{Corollary}{Corollaries}
\crefname{lemma}{Lemma}{Lemmas}
\crefname{equation}{Eq.}{Eq.}
\crefname{proposition}{Proposition}{Propositions}
\crefname{claim}{Claim}{Claims}
\crefname{remark}{Remark}{Remarks}
\crefname{observation}{Observation}{Observations}
\crefname{assumption}{Assumption}{Assumptions}
\crefname{template}{Template}{Template}
\crefname{definition}{Definition}{Definitions}
\crefname{appendix}{App.}{Apps.}
\crefname{algorithm}{Algorithm}{Algorithms}
\crefname{figure}{Fig.}{Fig.}
\crefname{table}{Table}{Tables}
\crefname{property}{Property}{Properties}
\crefname{line}{Line}{Lines}

\usepackage[most,skins,theorems]{tcolorbox}
\tcbset{
  aibox/.style={
    width=\linewidth,
    top=7pt,
    bottom=2pt,
    colback=blue!6!white,
    colframe=black,
    colbacktitle=black,
    enhanced,
    center,
    attach boxed title to top left={yshift=-0.1in,xshift=0.15in},
    boxed title style={boxrule=0pt,colframe=white,},
  }
}
\newtcolorbox{AIbox}[2][]{aibox,title=#2,#1}

\usepackage{soul}
\definecolor{highlightred}{RGB}{255, 204, 204}
\definecolor{highlightgreen}{RGB}{204, 255, 204}
\definecolor{frameblue}{RGB}{60, 100, 140}

\tcbset{
    llmstyle/.style={
        sharp corners,
        colback=white,
        boxrule=0.3mm,
        coltitle=black,
        fonttitle=\ttfamily\bfseries\small\color{black},
        colbacktitle=frameblue!50,
        enhanced,
        left=0.3mm, right=0.3mm, top=0.3mm, bottom=0.3mm,
    }
}

\definecolor{cas_blue}{HTML}{0014C7}
\definecolor{casit_red}{HTML}{C12A22}
\definecolor{liauto_blue}{HTML}{225450}
\newcommand{\pc}{{\color{casit_red}\boldsymbol{c}}}
\newcommand{\pu}{{\color{cas_blue}\boldsymbol{u}}}
\newcommand{\pl}{{\color{liauto_blue}\boldsymbol{l}}}

\newcommand{\ours}{\texttt{SATQuest}\xspace}
\definecolor{instance_yellow}{HTML}{554400}
\definecolor{problem_blue}{HTML}{000044}
\definecolor{question_green}{HTML}{004400}

\newcommand{\Psatdpsat}{\textcolor{problem_blue}{\texttt{SATDP-sat}}\xspace}
\newcommand{\Psatdpunsat}{\textcolor{problem_blue}{\texttt{SATDP-unsat}}\xspace}
\newcommand{\Psatdp}{\textcolor{problem_blue}{\texttt{SATDP}}\xspace}
\newcommand{\Psatsp}{\textcolor{problem_blue}{\texttt{SATSP}}\xspace}
\newcommand{\Pmaxsat}{\textcolor{problem_blue}{\texttt{MaxSAT}}\xspace}
\newcommand{\Pmcs}{\textcolor{problem_blue}{\texttt{MCS}}\xspace}
\newcommand{\Pmus}{\textcolor{problem_blue}{\texttt{MUS}}\xspace}

\newcommand{\Qdimacs}{\textcolor{question_green}{\texttt{DIMACS}}\xspace}
\newcommand{\Qmath}{\textcolor{question_green}{\texttt{Math}}\xspace}
\newcommand{\Qstory}{\textcolor{question_green}{\texttt{Story}}\xspace}
\newcommand{\Qdualstory}{\textcolor{question_green}{\texttt{DualStory}}\xspace}

\newcommand{\GPT}{\begin{ttfamily}GPT-4.1\end{ttfamily}\xspace}
\newcommand{\QwenSmall}{\begin{ttfamily}Qwen2.5-7B-Instruct\end{ttfamily}\xspace}
\newcommand{\QwenLarge}{\begin{ttfamily}Qwen2.5-32B-Instruct\end{ttfamily}\xspace}

\newcommand{\DSV}{\begin{ttfamily}DeepSeek-V3-0324\end{ttfamily}\xspace}
\newcommand{\oThreeMini}{\begin{ttfamily}o3-mini\end{ttfamily}\xspace}
\newcommand{\DSR}{\begin{ttfamily}DeepSeek-R1\end{ttfamily}\xspace}
\newcommand{\QwQ}{\begin{ttfamily}QwQ-32B\end{ttfamily}\xspace}
\newcommand{\DSRDistillQwenSmall}{\begin{ttfamily}DeepSeek-R1-Distill-Qwen-7B\end{ttfamily}\xspace}
\newcommand{\DSRDistillQwenLarge}{\begin{ttfamily}DeepSeek-R1-Distill-Qwen-32B\end{ttfamily}\xspace}

\newif\ifanonymous
\anonymousfalse

\ifanonymous
  \newcommand{\ourDataset}{\href{https://huggingface.co/datasets/ANON/SATQuest}{\huggingfacelogo~\texttt{ANON/SATQuest}}\xspace}
  \newcommand{\ourDatasetRFT}{\href{https://huggingface.co/datasets/ANON/SATQuest-RFT-3k}{\huggingfacelogo~\texttt{ANON/SATQuest-RFT-3k}}\xspace}

  \newcommand{\blindurl}[1]{\textcolor{red}{[Link hidden for anonymity]}}
  \newcommand{\blindhref}[2]{\textcolor{red}{[Link hidden for anonymity]}}
  \newcommand{\blindhreftiny}[2]{\textcolor{red}{\tiny{[Link hidden for anonymity]}}}
\else
  \newcommand{\ourDataset}{\href{https://huggingface.co/datasets/sdpkjc/SATQuest}{\huggingfacelogo~\texttt{sdpkjc/SATQuest}}\xspace}
  \newcommand{\ourDatasetRFT}{\href{https://huggingface.co/datasets/sdpkjc/SATQuest-RFT-3k}{\huggingfacelogo~\texttt{sdpkjc/SATQuest-RFT-3k}}\xspace}

  \newcommand{\blindurl}[1]{\url{#1}}
  \newcommand{\blindhref}[2]{\href{#1}{#2}}
  \newcommand{\blindhreftiny}[2]{\href{#1}{#2}}
\fi

\title{SATQuest: A Verifier for Logical Reasoning Evaluation and Reinforcement Fine-Tuning of LLMs}

\author{
Yanxiao Zhao\(^{\pc\hspace{.1em}\pu\hspace{.1em}\pl}\)\thanks{Work done during internship at \textit{Li Auto}.} \quad
Yaqian Li\(^{\pl}\) \quad
Zihao Bo\(^{\pl}\) \quad
Rinyoichi Takezoe\(^{\pl}\) \\
\textbf{Haojia Hui}\(^{\pl}\) \quad
\textbf{Mo Guang}\(^{\pl}\) \quad
\textbf{Lei Ren}\(^{\pl}\) \quad
\textbf{Xiaolin Qin}\(^{\pc\hspace{.1em}\pu}\)\thanks{\begin{tabular}[t]{@{}l@{\ }l@{}}
Corresponding author: & \texttt{qinxl2001@126.com},\\
& \texttt{longkaiwen@lixiang.com}
\end{tabular}} \quad
\textbf{Kaiwen Long}\(^{\pl}\)\footnotemark[2]  \\
\(^{\pc}\) Chengdu Institute of Computer Applications, Chinese Academy of Sciences \\
\(^{\pu}\) School of Computer Science and Technology, University of Chinese Academy of Sciences \\
\(^{\pl}\) Li Auto \\
}

\begin{document}
\maketitle

\begin{abstract}
Large language models (LLMs) exhibit strong general reasoning, yet the community lacks \emph{controllable, scalable, and verifiable} tools to analyze and improve these abilities.
We present \ours, a verifier that generates diverse SAT-based reasoning tasks directly from Conjunctive Normal Form (CNF) instances and checks answers objectively with PySAT.
\ours factorizes evaluation along three orthogonal dimensions—\emph{instance}, \emph{problem type}, and \emph{question format}—enabling fine-grained, multi-dimensional analysis and reinforcement fine-tuning.
Randomized CNF generation mitigates memorization and supports reproducible experiments.
Using \ours, we benchmark a range of open- and closed-weight LLMs and uncover persistent gaps in logical reasoning, particularly on higher-complexity tasks and in transfer beyond familiar mathematical notation to machine or narrative formats.
We further show that reinforcement fine-tuning with \ours rewards substantially boosts targeted performance and generalizes to larger instances, while cross-format robustness remains challenging.
Collectively, \ours provides verifier-backed infrastructure for controlled, scalable, and reproducible \emph{empirical research} on LLM logical reasoning and its training.
\end{abstract}

\begin{center}
    \faIcon{github}~\href{https://github.com/sdpkjc/SATQuest}{sdpkjc/SATQuest}
\end{center}

\section{Introduction}
\label{sec:introduction}

Large Language Models (LLMs) have demonstrated remarkable proficiency in general reasoning tasks, including complex problem-solving and code generation, with models like \texttt{o3-mini}\citep{openai2025o3mini}, \texttt{DeepSeek-R1}\citep{deepseekai2025deepseekr1incentivizingreasoningcapability}, and \texttt{QwQ-32B}\citep{qwq-32b-preview} excelling in programming, mathematics, and scientific question-answering\citep{NEURIPS2020_1457c0d6, openai2024gpt4technicalreport, wei2022chain}.
This advanced reasoning capability, a cornerstone for Artificial General Intelligence (AGI), is a critical indicator of models' deep understanding and generalization.

To further explore the reasoning capabilities of LLMs, we urgently need evaluation and training tools that are both controllable and scalable.
Fine-grained and reliable performance analysis requires systematic variable control, which is fundamental to empirical scientific research.
However, existing benchmarks and datasets have significant limitations in variable control, making it difficult to support multi-dimensional, systematic analysis and training experiments—hindering deeper understanding of reasoning mechanisms in LLMs.

Although benchmarks such as \textit{ZebraLogic}~\citep{lin2025zebralogicscalinglimitsllms} and \textit{Knights and Knaves}~\citep{xie2024memorization} have introduced controllable difficulty dimensions, their question types and formats remain narrow, supporting only the \textit{instance} dimension.
\textit{MATH}~\citep{hendrycks2021measuringmath} and \textit{LiveBench}~\citep{white2025livebench} provide rich content across mathematics and programming; however, they lack structured relationships and rely heavily on human judgment for difficulty labeling.
In comparison, general-purpose benchmarks like \textit{GPQA}~\citep{rein2024gpqa}, \textit{MMLU}~\citep{hendrycks2021measuring}, and \textit{Big-Bench}~\citep{srivastava2023beyond} offer broad coverage, yet suffer from issues such as data leakage, lack of continuity, and insufficient support for multi-dimensional controlled analysis.

In this work, we aim to construct a multi-dimensionally controllable and scalable verifier to support the evaluation and training of LLM reasoning capabilities, enabling the tracking of progress and discovery of limitations in LLM reasoning.
Our main contributions are as follows:


(1) \textbf{CNF-native, controllable verifier.} We present \ours, which generates SAT-based reasoning tasks directly from CNF and organizes them along three orthogonal axes—\emph{instance}, \emph{problem type}, and \emph{question format}. Randomized CNF construction, binary-string targets, and objective PySAT checking enable leakage-resistant, verifiable evaluation and training; we also release \ourDataset for reproducibility.

(2) \textbf{Multi-dimensional evaluation.} Using \ours, we provide a comprehensive study of open/closed-weight LLMs across all three axes, revealing strong cross-format sensitivity beyond \Qmath, accuracy degradation with solver-measured complexity, and a systematic shift toward hallucination (fabricated execution and invented shortcuts) under high complexity (\cref{app:hallucination-audit}).

(3) \textbf{Verifier-driven Reinforcement Fine-tuning.} We implement reinforcement fine-tuning with \ours rewards, yielding targeted gains and longer reasoning chains. Improvements transfer to larger instances and exhibit \emph{asymmetric} cross-task generalization, while cross-format robustness remains limited.

\section{SATQuest Challenge Design}
\label{sec:challenge}

\paragraph{Overview.}
\ours is a systematic verifier engineered to comprehensively evaluate and enhance the logical reasoning capabilities of LLMs.
Its primary goal is to offer a framework for fine-grained analysis, providing deeper insights into the strengths and limitations of LLMs in logical deduction.
\ours is not designed to train LLMs as general-purpose solvers or to make them surpass specialized symbolic solvers in speed or accuracy.

To achieve its objectives, \ours automatically generates a variety of logical reasoning tasks directly derived from CNF instances.
These tasks are meticulously organized along three orthogonal dimensions: \textit{instance}, \textit{problem type}, and \textit{question format}, each targeting distinct aspects of logical reasoning.
This multi-dimensional structure creates a comprehensive and controllable challenge space suitable for nuanced evaluation and effective reinforcement fine-tuning.
CNF instances are utilized as the foundational elements due to their formal clarity, their established role as a standard representation in propositional logic, and their inherent compatibility with established SAT solvers for objective answer verification, making them an ideal medium for systematically probing the logical capabilities of LLMs.

\paragraph{Data Generation.}
\ours supports evaluation using any CNF instance dataset stored in the standard \Qdimacs format.
For reproducibility, we generated two CNF datasets using the procedure outlined in \cref{alg:gen-cnf}:

\begin{itemize}[leftmargin=1em]
\item \ourDataset:
This dataset is generated for evaluation purposes (\cref{sec:evaluations}), specifically to assess logical reasoning across varying instance scales and difficulties.
It consists of randomly generated CNF instances with \(n \in [3, 16]\) variables and a fixed clause-to-variable ratio resulting in \(m = 4n\) clauses.
For each \((n, m)\) configuration, \(10\) CNF instance pairs (one satisfiable and one unsatisfiable) were generated, resulting in a total of \(140\) CNF pairs.
The two CNF instances share the same number of literals and nearly identical CNF structures.

\item \ourDatasetRFT:
This dataset is generated for reinforcement fine-tuning (\cref{sec:rft}). It consists of CNF instances with \(n \in [3, 8]\) variables and clause counts \(m\) determined by varying the clause-to-variable ratio from \(2.1\) to \(4.0\) in increments of \(0.1\) (i.e., \(m\) ranges from \(2.1n\) to \(4.0n\)).
For each \((n, m)\) configuration, \(25\) CNF instance pairs were generated, resulting in a total of \(3,000\) CNF pairs.
\end{itemize}

\paragraph{Challenge Dimensions 1: \textcolor{instance_yellow}{\textit{Instance}}.}
We categorize instances by their \textit{scale} (number of variables \(n\), clauses \(m\), and literals) and inherent \textit{difficulty}.
Computational \textit{difficulty} is assessed using SAT solver statistics: \textit{decisions}, indicating search breadth; \textit{conflicts}, reflecting constraint-driven backtracking; and \textit{propagations}, quantifying chained logical implications.
These structural and solver-derived metrics provide a multi-faceted characterization of an instance's combinatorial complexity, where higher values generally correspond to more challenging problems.

\paragraph{Challenge Dimensions 2: \textcolor{problem_blue}{\textit{Problem Type}}.}
Let $\alpha : X \to \{0,1\}$ be an assignment for $X=\{x_1,\dots,x_n\}$, and let $F=\bigwedge_{i=1}^m C_i$ be a CNF formula.
Each $C_i$ is a clause of the form $C_i = \ell_{i,1} \lor \ell_{i,2} \lor \cdots \lor \ell_{i,k_i}$, where $\ell_{i,j} \in X \cup \{\lnot x | x \in X\}$.
We define five fundamental SAT-based problems:

\begin{itemize}[leftmargin=1em]
  \item \Psatdp~(SAT Decision Problem): Determine whether \(F\) is satisfiable:
  \[
    \texttt{SATDP}(F) =
    \begin{cases}
      1, & \exists\,\alpha \text{ s.t. } F(\alpha)=1,\\
      0, & \text{otherwise}.
    \end{cases}
  \]
  Tests the fundamental ability to determine logical consistency.
  \item \Psatsp~(SAT Search Problem): If \(F\) is satisfiable, find an assignment \(\alpha\):
  \[
    \alpha \text{ s.t. } F(\alpha)=1.
  \]
  Probes constructive reasoning by requiring the generation of a satisfying assignment.
  \item \Pmaxsat~(Maximum Satisfiability): Find the assignment \(\alpha^*\) that maximizes the number of satisfied clauses:
  \[
    \alpha^* = \arg\max_{\alpha} \sum_{i=1}^m \mathbf{1}[C_i(\alpha)=1].
  \]
  Evaluates optimization skills when maximizing clause satisfaction under conflicting constraints.
  \item \Pmcs~(Minimal Correction Subset): For an unsatisfiable \(F\), find a minimal set \(S \subseteq \{1,\dots,m\}\) whose removal yields a satisfiable formula. Here \(S\) indexes the removed clauses, so satisfiability is checked on the remaining clauses:
  \begin{multline*}
    \texttt{SATDP}(\bigwedge_{i \notin S} C_i) = 1 \bigwedge \\
    \forall\, S^\prime \subset S,\ \texttt{SATDP}(\bigwedge_{i \notin S^\prime} C_i) = 0 .
  \end{multline*}
  Tests diagnostic reasoning through the identification of minimal corrections for unsatisfiability.
  \item \Pmus~(Minimal Unsatisfiable Subset): For an unsatisfiable \(F\), find a minimal unsatisfiable core \(S \subseteq \{1,\dots,m\}\):
  \begin{multline*}
    \texttt{SATDP}(\bigwedge_{i \in S} C_i) = 0 \bigwedge \\
    \forall\, S^\prime \subset S,\ \texttt{SATDP}(\bigwedge_{i \in S^\prime} C_i) = 1 .
  \end{multline*}
  Probes diagnostic reasoning by localizing minimal sources of logical inconsistency.
\end{itemize}

Each CNF instance pair consists of one satisfiable and one unsatisfiable formula.
The satisfiable instance is used for \Psatsp, while the unsatisfiable instance is used for \Pmaxsat, \Pmcs, and \Pmus. For \Psatdp, both instances are evaluated, forming two sub-tasks (\Psatdpsat and \Psatdpunsat).
A response to \Psatdp is considered correct only if both sub-tasks are answered correctly, thereby discouraging random guessing.
These problems progressively challenge LLM reasoning capabilities, from foundational deduction and solution construction (\Psatdp/\Psatsp), to constrained optimization (\Pmaxsat), and finally to minimal cause identification and correction (\Pmcs/\Pmus).

\paragraph{Challenge Dimensions 3: \textcolor{question_green}{\textit{Question Format}}.}
Recognizing that the presentation of a problem can significantly impact an LLM's reasoning process, this dimension introduces four logically equivalent representational formats for each CNF instance.
This variation aims to test different reasoning skills and reduce reliance on superficial pattern matching.

\begin{itemize}[leftmargin=1em]
  \item \Qmath~(mathematical notation): Uses \(\land\), \(\lor\), and \(\lnot\) to represent logic formulas. Balances between formality and readability.\\
  \emph{Example:} \(x_1 \lor \lnot x_2 \lor x_3\)

  \item \Qdimacs~(machine format): A minimal, line-based format for representing Boolean formulas in CNF, the standard input for many SAT symbolic solvers.\\
  \emph{Example:}\vspace{-1em}
\begin{verbatim}
        p cnf 3 1
        1 -2 3 0
\end{verbatim}
\vspace{-1em}

  \item \Qstory~(OR semantics, cookie day scenario): Wraps clauses as friendly narratives—"Alice is happy if..."—to test LLMs' ability to ground disjunctions in natural language.\\
        \emph{Example:} \emph{"Alice will be happy if she gets \underline{crunchy choco} (\(x_1\)), \underline{chewy vanilla} (\(\lnot x_2\)), or \underline{crunchy peanut} (\(x_3\)).''}

  \item \Qdualstory~(AND semantics, cookie day scenario): Presents the negated form—"Alice will be unhappy only if..." —turning OR into AND and requiring semantic tracking.\\
        \emph{Example:} \emph{"Alice will be unhappy only if she is served \underline{crunchy choco} (\(x_1\)), \underline{chewy vanilla} (\(\lnot x_2\)), and \underline{crunchy peanut} (\(x_3\)).''}
\end{itemize}

\Qmath is common in training data and accessible to math-tuned LLMs, whereas \Qdimacs is a compact, noise-free, machine-readable format that tests a model's ability to interpret raw clause structures.
We adopt it specifically to evaluate whether LLMs can interpret raw clause structures without relying on mathematical training.
\Qstory and \Qdualstory introduce narrative elements that add informational noise and require the model to translate natural-language logical structures into formal logic before reasoning.
The deterministic CNF$\to$narrative conversion maps variables to cookie names (from a predefined pool), positive/negative literals to ``crunchy''/``chewy'' variants, and clauses to individual friend preferences; the complete mapping procedure with pseudocode is given in \cref{app:cnf-to-narrative}.

\begin{figure}[h]
    \centering
    \includegraphics[width=0.9\linewidth]{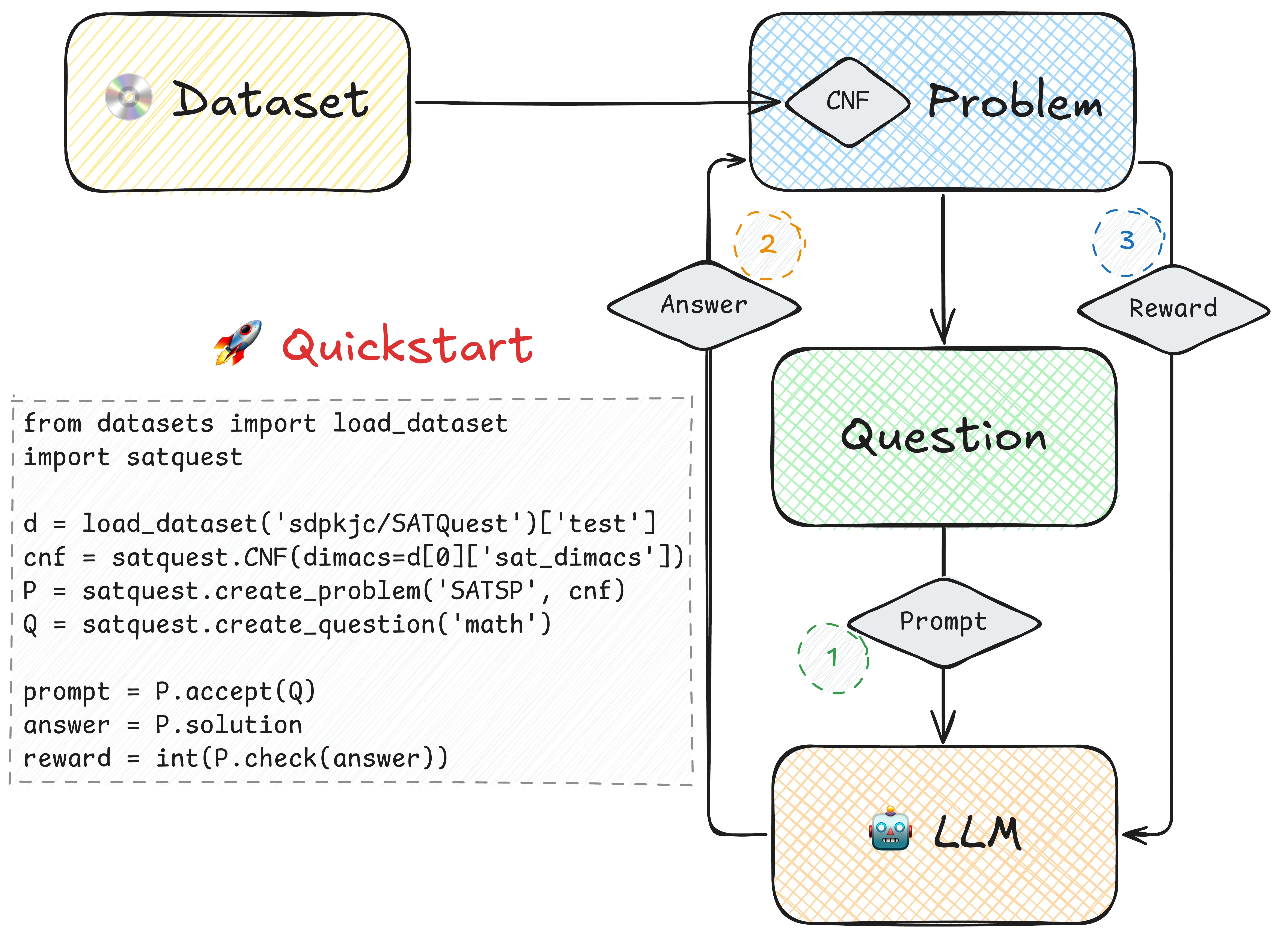}
    \caption{\ours Pipeline and Quickstart.}
    \label{fig:pipeline}
    \vspace{-1.5em}
\end{figure}

\paragraph{Answer Verification.}
Prior logical reasoning benchmarks (e.g., \citep{lin2025zebralogicscalinglimitsllms, xie2024memorization, mondorf-plank-2024-liar}) typically require LLMs to produce \texttt{JSON} outputs, choose from preset options, or use complex extraction—methods that hinder automation and scalability despite reduced manual overhead.
Moreover, multiple-choice formats can significantly alter the true difficulty of problems, as the complexity of \Psatsp differs fundamentally from SAT verification.
\ours uses a unified binary-string output format—1 bit for \Psatdp, n bits for \Psatsp/\Pmaxsat, and m bits for \Pmcs/\Pmus—to ensure consistency across tasks and to control for variation in the output schema.
These strings are extracted using regex and checked against CNF constraints via PySAT, allowing for multiple valid answers.
A known limitation is that long binary outputs may challenge smaller models' ability to adhere to the expected format.
To shed light on this, \cref{fig:heatmaps_format} presents format correctness statistics for mainstream models evaluated.
For the complete prompt and detailed output format instructions, refer to \cref{app:prompt-details}.
The \ours Pipeline and Quickstart are illustrated in \cref{fig:pipeline}.

\section{Evaluation}
\label{sec:evaluations}

\paragraph{Overview.}
We conduct a comprehensive evaluation of the logical reasoning performance of various LLMs using the \ours benchmark.
The analysis spans different \textit{instance}, \textit{problem types}, and \textit{question formats}, enabling fine-grained, multi-dimensional insights.
We begin with the overall benchmark results, followed by detailed analyses along each dimension.

\paragraph{Setup.}
We evaluate a diverse set of state-of-the-art open-weight and closed-weight LLMs, including vanilla models, reasoning models, and distilled variants.
The evaluation uses the \ourDataset dataset, comprising \(140\) CNF instance pairs categorized by scale (\(n \in [3,16]\), clauses \(m=4n\)).
Each CNF pair yields tasks across five logical reasoning types (\Psatdp, \Psatsp, \Pmaxsat, \Pmcs, \Pmus) and four question formats (\Qmath, \Qdimacs, \Qstory, \Qdualstory), resulting in \(20\) evaluations per CNF pair.
For detailed evaluation configurations and parameters, see \cref{app:evaluation-configs}.

\begin{figure}[ht]
    \centering
    \includegraphics[width=1.0\linewidth]{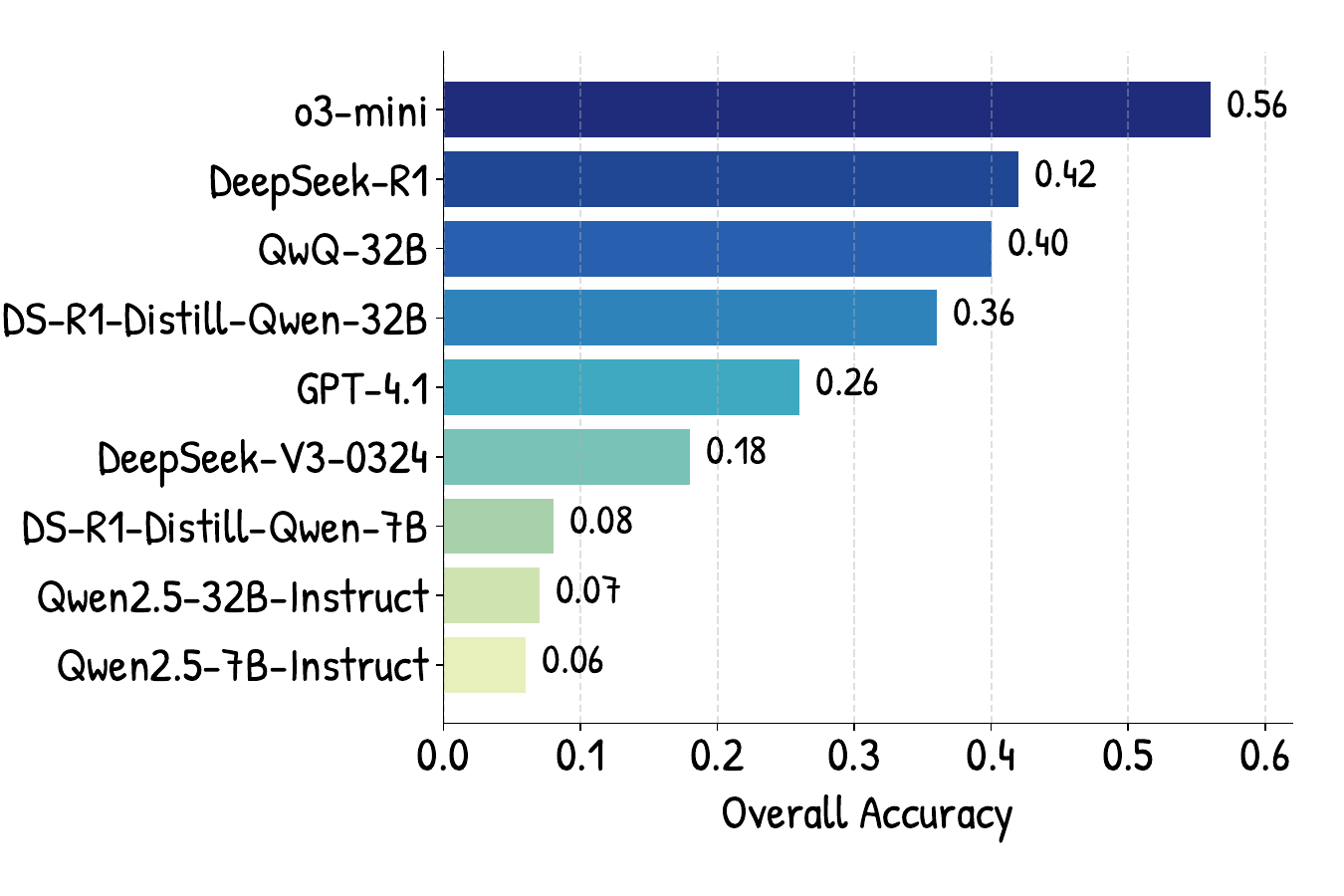}
    \vspace{-2em}
    \caption{Overall accuracy of evaluated LLMs on the \ours benchmark, averaged across all \textit{problem types} and \textit{question formats}.}\label{fig:overall_accuracy}
    \vspace{-1em}
\end{figure}

\paragraph{Overall Results.}
\cref{fig:overall_accuracy} shows model accuracies on \ours. \oThreeMini leads with \(0.56\) accuracy, followed by \DSR (\(0.42\)), \QwQ (\(0.40\)), and \DSRDistillQwenLarge(\(0.39\)), indicating that reasoning-enhanced models outperform vanilla LLMs.
Large vanilla models like \GPT (\(0.38\)) and \DSV (\(0.36\)) perform competitively despite lacking explicit reasoning training.
In contrast, smaller vanilla models (e.g., \QwenSmall) achieve below \(0.1\) accuracy, revealing limited reasoning capabilities.

These results point to two notable trends.
First, reasoning models consistently outperform vanilla counterparts, particularly on more complex tasks.
Second, the modest overall accuracy across models reflects the challenging nature of \ours and its effectiveness in distinguishing reasoning capabilities.
Moreover, the model ranking induced by \ours is broadly consistent with that of other recent reasoning benchmarks such as \textit{GPQA}~\citep{rein2024gpqa} and \textit{ZebraLogic}~\citep{lin2025zebralogicscalinglimitsllms}, lending confidence that \ours reliably differentiates reasoning capability within the evaluated LLM population.

\begin{figure}[ht]
    \centering
    \includegraphics[width=0.9\linewidth]{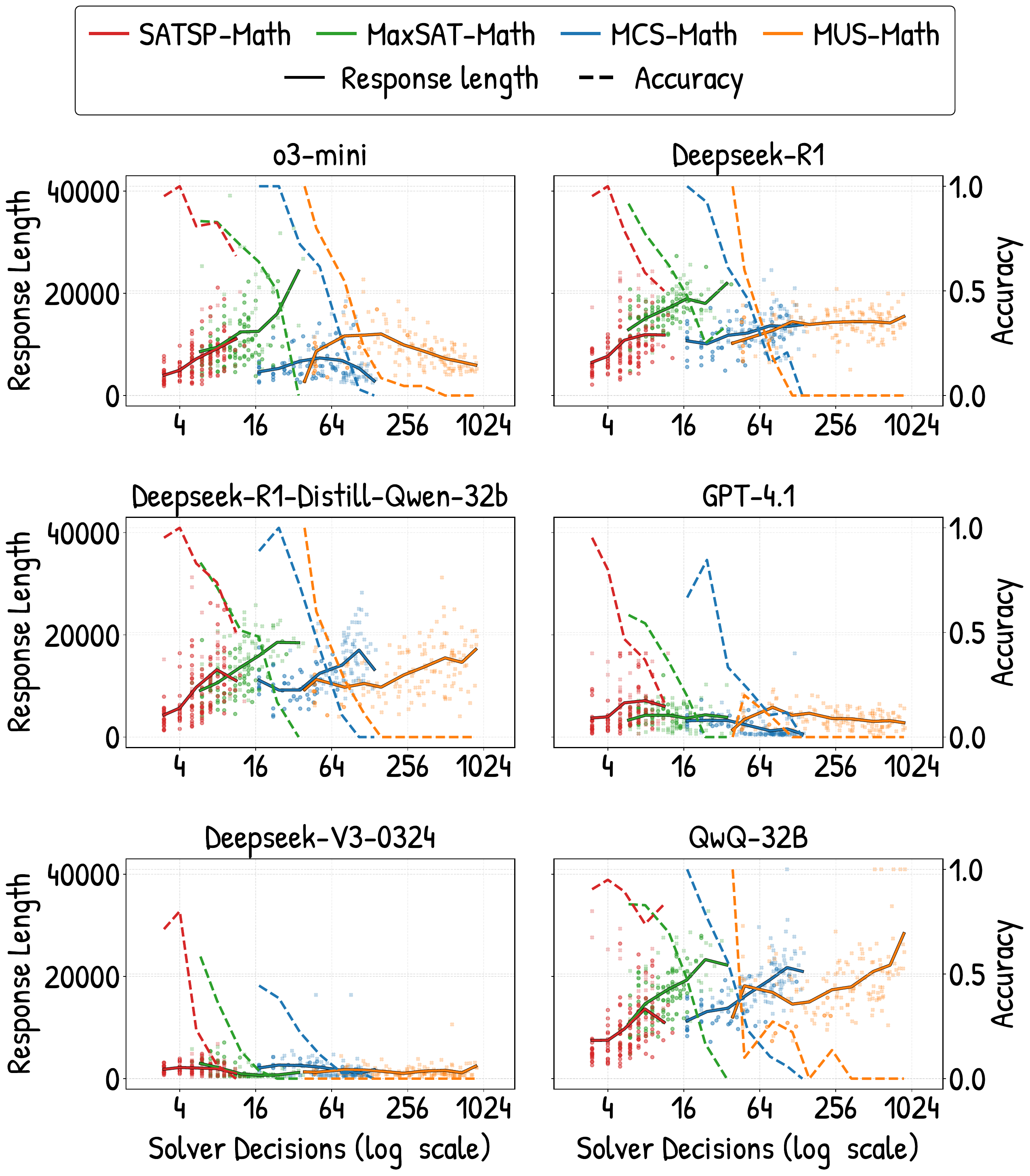}
    \vspace{-0.5em}
    \caption{Accuracy (dashed lines, left axis) and response length (solid lines, right axis) vs. instance complexity (solver decisions) for \Psatsp-\Qmath, \Pmaxsat-\Qmath, \Pmcs-\Qmath and \Pmus-\Qmath tasks.}\label{fig:scale_difficulty}
    \vspace{-1.5em}
\end{figure}

\paragraph{Analysis by \textcolor{instance_yellow}{\textit{Instance}}.}
We examine how model performance scales with instance complexity, measured using the number of decisions made by established SAT solvers (Glucose 4.1~\citep{doi:10.1142/S0218213018400018} for \Psatdp/\Psatsp, RC2~\citep{doi:10.3233/SAT190116} for \Pmaxsat, LBX~\citep{10.5555/2832415.2832523} for \Pmcs, and MUSX~\citep{10.1109/ISMVL.2010.11} for \Pmus).
\cref{fig:scale_difficulty} visualizes model accuracy and response length against this complexity metric.
A consistent trend across tasks is that as instance complexity increases (more solver decisions), model accuracy tends to decline, while response length generally increases.
Notably, we observe a concerning hallucination phenomenon when models encounter highly complex instances: they often fabricate solver calls or invent simplified reasoning paths rather than engaging with the full logical complexity of the problem.
This hallucination is particularly evident for \texttt{o3-mini} on \Pmcs-\Qmath and \Pmus-\Qmath tasks, where response length actually decreases at high complexity, indicating the model abandons complete reasoning in favor of hallucinated shortcuts.
Overall, all models exhibit reduced accuracy on larger and more difficult instances. Top-performing models like \texttt{o3-mini} and \texttt{DeepSeek-R1} show more gradual degradation, indicating better scalability, whereas less capable models experience a sharp performance drop, often failing completely on moderately complex instances and resorting to increasingly severe hallucinations when faced with logical complexity beyond their reasoning capacity.

\begin{figure}[ht]
    \centering
    \includegraphics[width=1\linewidth]{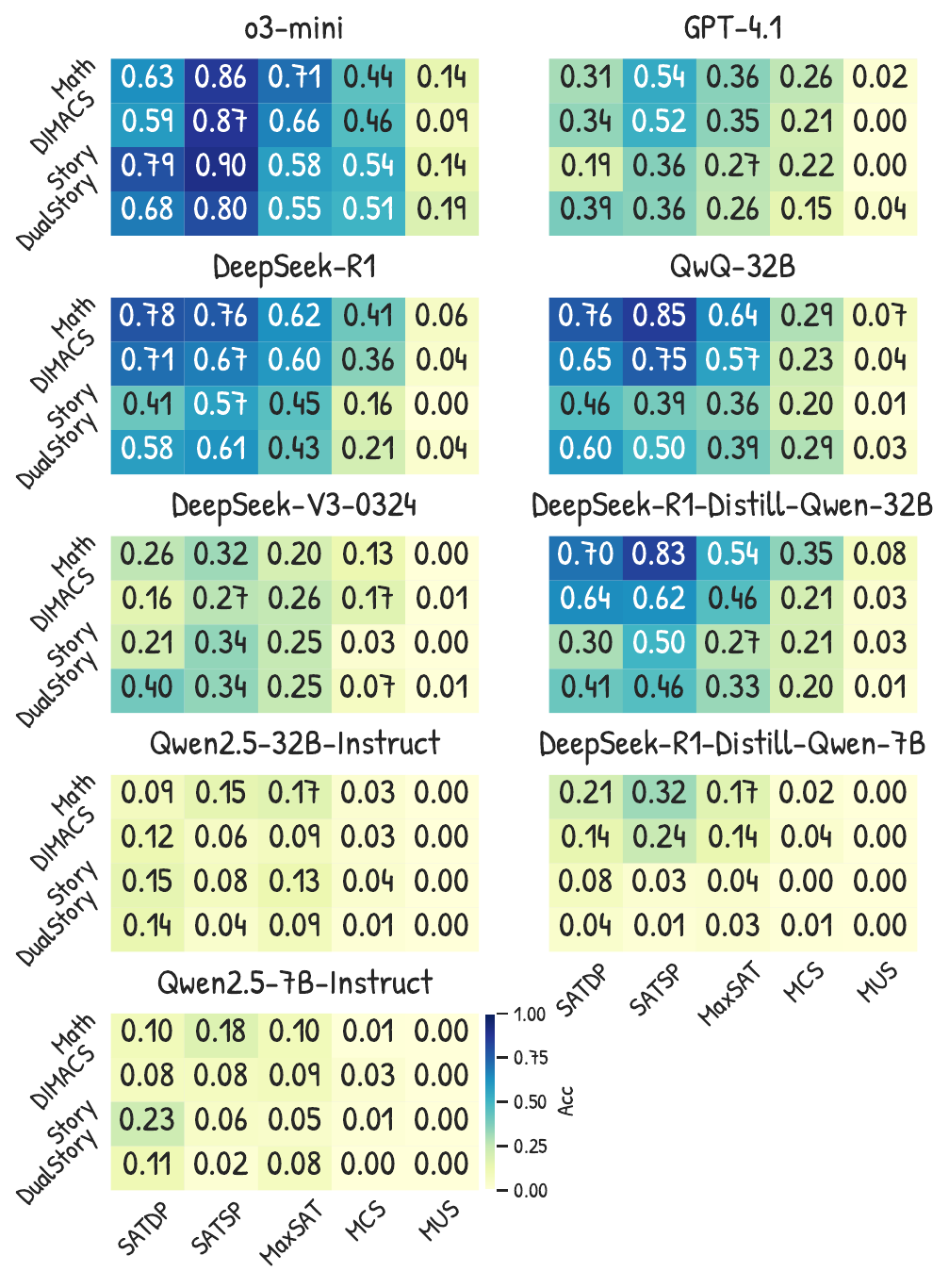}
    \vspace{-2em}
    \caption{Accuracy heatmaps from a single run for each model, showing performance by \textit{problem type} (columns) and \textit{question format} (rows).}\label{fig:heatmaps}
    \vspace{-1.5em}
\end{figure}

\paragraph{Analysis by \textcolor{problem_blue}{\textit{Problem Type}}.}
We analyze performance variations across the five distinct SAT-based problem types: \Psatdp, \Psatsp, \Pmaxsat, \Pmcs, and \Pmus.
The heatmaps in \cref{fig:heatmaps} reveal performance differences across these tasks.
Consistently across models, performance tends to be highest on \Psatdp (especially \Psatdpsat) and \Psatsp, and lowest on \Pmcs and \Pmus, forming a clear difficulty hierarchy.
Models generally handle the basic decision (\Psatdp) and search (\Psatsp) tasks better than tasks requiring optimization or diagnosis, though performance on \Psatsp is often lower than \Psatdp.
Performance on the optimization task \Pmaxsat typically sits between the basic tasks and the diagnostic tasks.
The diagnostic tasks, \Pmcs and \Pmus, which require identifying minimal subsets, prove particularly challenging, with accuracy dropping significantly for almost all models.
While top models like \texttt{o3-mini} maintain some capability even on harder tasks, the gap between task types is pronounced across the board.
This performance stratification aligns with the solver complexity shown in \cref{fig:scale_difficulty} (measured by solver decisions), suggesting LLMs struggle progressively more with tasks demanding global optimization, minimality constraints, and diagnostic reasoning over combinatorial spaces.
Overall, the results highlight LLM limitations in tackling the full spectrum of logical reasoning challenges represented by these diverse SAT-based tasks.

\paragraph{Analysis by \textcolor{question_green}{\textit{Question Format}}.}
\label{sec:eval-question-format}
The way a logical problem is presented can significantly affect an LLM's ability to solve it. We analyze this impact by evaluating performance across four distinct question formats: \Qmath, \Qdimacs, \Qstory, and \Qdualstory.
\cref{fig:heatmaps} illustrates how accuracy varies across these formats.
All models perform best in the \Qmath format, generally achieving their highest accuracy, followed by \Qdimacs, with \Qstory and \Qdualstory formats yielding the lowest accuracy.
\texttt{o3-mini} demonstrates relatively stable performance across the four formats, indicating strong reasoning robustness regardless of presentation style.
However, other open-weight reasoning models like \texttt{DeepSeek-R1} and \texttt{QwQ-32B}, while performing well in the \Qmath format, exhibit a significant drop in accuracy in other formats, suggesting higher sensitivity to the presentation style.

It is noteworthy that the \Qstory and \Qdualstory problems introduce narrative elements, adding informational noise and requiring the model to translate the natural language logical structure into formal logic before reasoning.
The increased difficulty and subsequent lower accuracy are thus expected.
However, the \Qdimacs format is structurally similar to \Qmath, contains no redundant information, and has higher information density.
Despite this, open-weight reasoning models still show a marked decrease in accuracy compared to \Qmath (e.g., \texttt{DeepSeek-R1} and \texttt{QwQ-32B} accuracy dropped by 9\% and 10\% respectively on \Psatsp-\Qdimacs compared to \Psatsp-\Qmath).

Through case studies presented in \cref{app:case-studies}, we observe that \texttt{DeepSeek-R1} and \texttt{QwQ-32B} often attempt to reason directly within the \Qdimacs format rather than translating it into formal mathematical notation.
This approach involves working with the raw \Qdimacs clauses, which requires tracking multiple variable assignments simultaneously across numerous constraints.
The models frequently make errors when attempting to verify clause satisfaction or when determining the implications of specific variable assignments, particularly misinterpreting the disjunctive nature of clauses or conflating the semantic meaning of positive and negative literals.
This direct approach appears to lead to a higher error rate during the reasoning process, as the models struggle to maintain consistency across the complex network of logical constraints represented in the \Qdimacs format.

Conversely, vanilla models like \GPT and \DSV, although performing worse overall, show relatively balanced performance across different formats, indicating lower format sensitivity.
These models sometimes employ structured thinking approaches by first translating \Qdimacs inputs into the \Qmath format before proceeding with reasoning, or by introducing meaningful symbolic notation during their reasoning process, which appears to enhance reasoning stability.
The shorter reasoning chains produced by vanilla models may also contribute to their format robustness, as briefer deductions have fewer opportunities for errors to accumulate.
This contrasts with \DSR and \QwQ, whose struggles outside the \Qmath format seem to stem from relying more on potentially error-prone direct reasoning or trial-and-error within unfamiliar formats, rather than employing systematic format translation or structured analysis.

Achieving AGI likely requires LLMs to reason effectively across diverse formats, enabling the integration of knowledge from different domains and fostering more powerful, generalized reasoning capabilities.
\ours thus serves as a valuable benchmark for assessing LLMs' adaptability and robustness in logical reasoning across various presentation styles.

\section{Reinforcement Fine-Tuning}
\label{sec:rft}

\paragraph{Overview.}
As demonstrated by the evaluation in \cref{sec:evaluations}, current LLMs exhibit significant limitations in logical reasoning, with notable deficiencies in generalization across \textit{instance}, \textit{problem type}, and \textit{question format}.
This section explores avenues for enhancing LLM logical reasoning capabilities through Reinforcement Fine-Tuning (RFT), directly utilizing reward signals from the \ours verifier.
Our investigation particularly focuses on two aspects:
First, we assess whether \ours-driven RFT can stimulate LLMs to construct longer reasoning chains, thereby fostering deeper logical deduction.
Second, we delve into the relationship between RFT and the generalization deficiencies identified in \cref{sec:evaluations}, aiming to elucidate the specific effects and potential bottlenecks of RFT in improving cross-task and cross-format generalization.

\begin{figure*}[ht]
    \centering
    \includegraphics[width=0.9\linewidth]{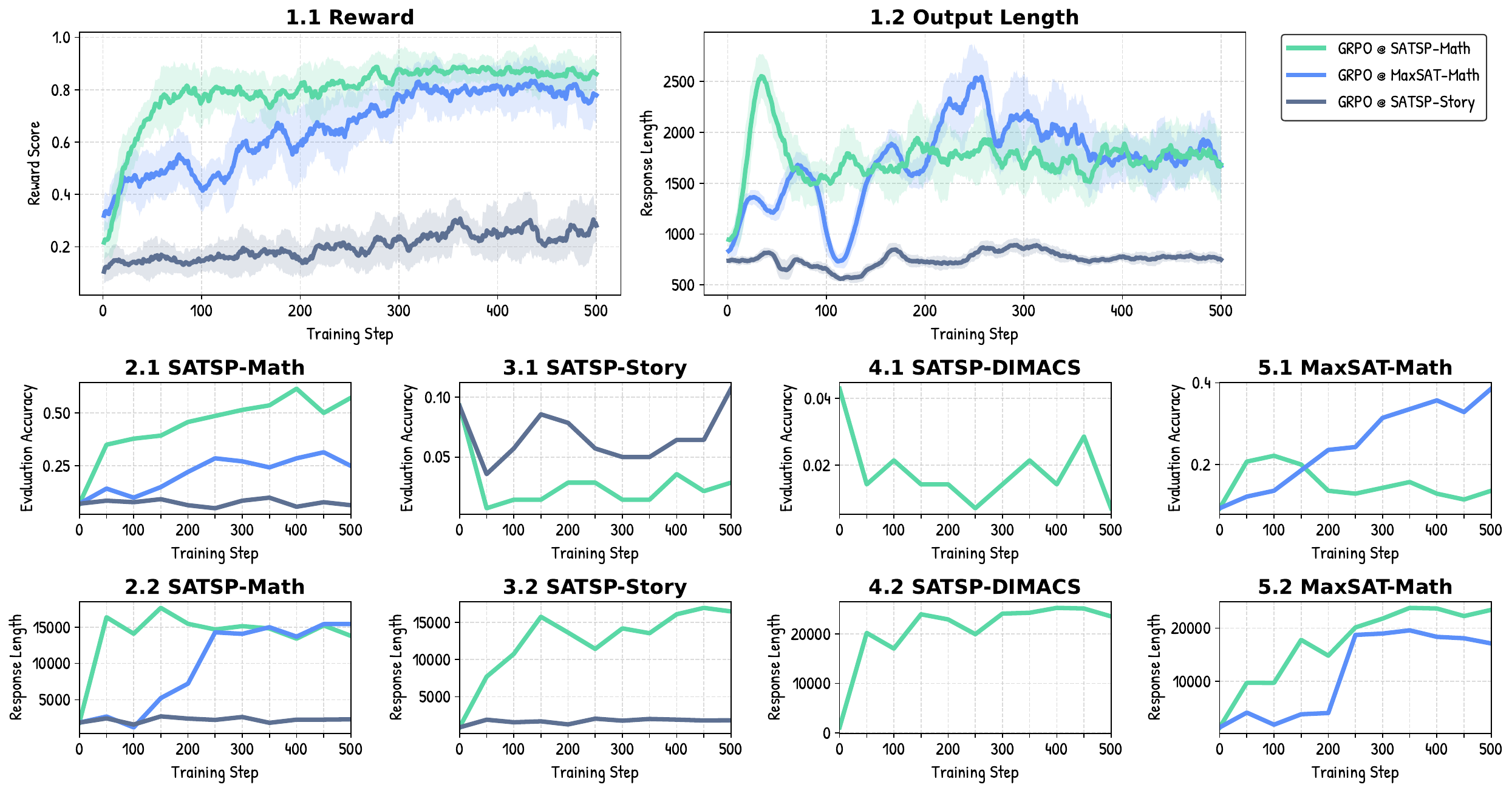}
    \vspace{-0.5em}
    \caption{GRPO fine-tuning using \ours rewards. Training dynamics (Top: reward, response length) and evaluation performance (Bottom: accuracy, response length on target and generalization tasks) vs. training steps for models fine-tuned on \Psatsp-\Qmath, \Pmaxsat-\Qmath, and \Psatsp-\Qstory. Solid lines show the mean; shaded regions denote $\pm1$ standard deviation.}
    \label{fig:rft_training_and_eval}
    \vspace{-1.5em}
\end{figure*}

\paragraph{Setup.}
We select the \texttt{Qwen2.5-7B-Instruct} model as the baseline, utilizing the Group Relative Policy Optimization (GRPO) algorithm~\citep{shao2024deepseekmath} during fine-tuning.
This baseline model is nearly a blank slate, a small vanilla model that only shows marginal performance on \Psatsp-\Qmath.
The training leverages the \ourDatasetRFT dataset, comprising \(3,000\) CNF instance pairs with \(n \in [3, 8]\) variables and clause-to-variable ratios ranging from \(2.1\) to \(4.0\).
We train three distinct models, each focusing on a specific \textit{problem type} and \textit{question format} combination: \Psatsp-\Qmath, \Psatsp-\Qstory, \Pmaxsat-\Qmath.

Our implementation is based on the TRL library~\citep{vonwerra2022trl}, adopting the GRPO objective as described in~\citep{shao2024deepseekmath}.
The prompt template follows the structure proposed in~\citep{deepseekai2025deepseekr1incentivizingreasoningcapability}.
The reward function is designed using the \ours verifier, assigning a reward of \(1.0\) for correct answers and \(0.0\) for incorrect ones.
Additionally, two format correctness rewards, detailed in \cref{app:format-reward}, are incorporated with weights of \(0.05\) each, complementing the primary reward weight of \(1.0\).
For training we set \(max\_prompt\_length\) to \(2048\) and \(max\_completion\_length\) to \(8192\), for evaluation we set \(max\_prompt\_length + max\_completion\_length\) to \(32768\).
The training and evaluation parameters are detailed in \cref{app:training-params}.

\paragraph{Training Dynamics.}
Training curves (\cref{fig:rft_training_and_eval}, top row) reveal that models trained on \Qmath-based tasks (\Psatsp-\Qmath and \Pmaxsat-\Qmath) achieve higher rewards and generate longer responses than those trained on \Psatsp-\Qstory.
The \Qmath format appears to better facilitate extended reasoning chains that receive positive reinforcement from the verifier.
Our \ours verifier effectively stimulates extended reasoning development within few training steps, especially with the mathematical format.
Response length curves show distinct patterns—rapid initial growth as models learn longer reasoning chains, temporary decline when adapting to format constraints at the training response limit (\(8192\) tokens), followed by stabilization.

Training on \Psatsp-\Qstory proved less effective, largely due to the baseline model's weak narrative reasoning abilities.
While \textit{Logic-RL}~\citep{xie2025logicrlunleashingllmreasoning} has successfully stimulated narrative reasoning, our tasks involve substantially higher complexity and scale.
Convergence efficiency correlates with task complexity and initial model capabilities, though these factors require further investigation to fully separate.

\paragraph{Generalization Across \textcolor{instance_yellow}{\textit{Instance}}.}
We observe positive generalization concerning problem complexity.
Models fine-tuned on \Psatsp-\Qmath and \Pmaxsat-\Qmath demonstrated improved accuracy when evaluated on the corresponding tasks within the evaluation set (\cref{fig:rft_training_and_eval}, subplots 2.1, 5.1).
Crucially, these evaluation instances involved larger scales ($n > 8$) than those used during training ($n \in [3, 8]$), indicating that the learned reasoning skills generalize to more complex instances within the same problem and format.

\begin{table*}[!t]
    \centering
    \begin{tabular}{l c c}
        \toprule
        \textbf{Generalization Type} & \textbf{Positive Transfer?} & \textbf{Observation} \\
        \midrule
        Across Instances     & \cmark                 & Improved accuracy on larger unseen CNF instances. \\
        Across Problem Types & \cmark\ (Asymmetric)   & \Pmaxsat $\rightarrow$ \Psatsp transfer stronger. \\
        Across Question Formats       & \xmark                 & \Qmath $\rightarrow$ \Qdimacs/\Qstory weak transfer. \\
        \bottomrule
    \end{tabular}
    \caption{Summary of generalization findings across \ours\ dimensions.}
    \label{tab:generalization-findings}
    \vspace{-1em}
\end{table*}

\paragraph{Generalization Across \textcolor{problem_blue}{\textit{Problem Types}}.}
Our results reveal an interesting asymmetry in cross-problem generalization.
Fine-tuning on the more complex \Pmaxsat-\Qmath task led to performance improvements not only on \Pmaxsat-\Qmath itself but also conferred benefits to the simpler \Psatsp-\Qmath task.
However, the model trained solely on \Psatsp-\Qmath did not show a corresponding improvement on \Pmaxsat-\Qmath (compare improvements patterns in \cref{fig:rft_training_and_eval}, bottom row).
This suggests that the reasoning capabilities required for \Pmaxsat may encompass those needed for \Psatsp.
Strategically, this implies that training on more complex and diverse logical problems could be more effective for fostering robust reasoning skills that generalize to simpler, related problems.

\paragraph{Generalization Across \textcolor{question_green}{\textit{Question Formats}}.}
Cross-format generalization remains notably difficult. The model fine-tuned on \Psatsp-\Qmath shows minimal improvement when evaluated on other formats such as \Psatsp-\Qstory (\cref{fig:rft_training_and_eval}, subplot 3.1) and even on the structurally similar \Psatsp-\Qdimacs task (subplot 4.1).
This suggests that reasoning capabilities acquired in the \Qmath format do not readily transfer to logically equivalent tasks presented in alternative formats, whether narrative or machine-readable.
Further analysis of failure cases reveals that, after \Psatsp-\Qmath fine-tuning, the model tends to generate verbose but flawed reasoning when confronted with \Qdimacs inputs—mirroring the issues described in \cref{sec:eval-question-format,app:case-studies}.
The model appears to have overfitted to a specific \Qmath-style reasoning pattern, at the expense of its initial structured thinking ability, and fails to effectively translate \Qdimacs representations into a suitable reasoning form.
These findings indicate that small-scale RFT may not be sufficient to overcome format generalization barriers, and that the performance discrepancies across formats observed in \cref{sec:evaluations} may partially stem from the limitations of format-specific fine-tuning itself.

\section{Related Work}
\label{sec:related-work}

Researchers have developed numerous benchmarks to evaluate LLMs' capabilities.
Popular evaluations like \citep{rein2024gpqa, hendrycks2021measuringmath, hendrycks2021measuring, srivastava2023beyond} comprehensively assess LLMs but suffer from data leakage and lack of continuity, often being solved by advanced models within 18 months of introduction.
While newer benchmarks \citep{suzgun-etal-2023-challenging, kazemi2025bigbenchextrahard, pteam2025supergpqascalingllmevaluation, gema-etal-2025-done, wang2024mmlupro, glazer2024frontiermathbenchmarkevaluatingadvanced} offer improvements, core issues remain.
\citet{white2025livebench} introduced dynamic question banks and automatic scoring but still relies on manual difficulty annotation without multi-dimensional analysis controls.
These evaluations primarily assess capabilities rather than providing insights into internal mechanisms.
Some studies explore more nuanced approaches: \citet{lin2025zebralogicscalinglimitsllms, xie2024memorization} use formalized templates with controllable difficulty dimensions for finer-grained analysis.
\citet{xie2025logicrlunleashingllmreasoning} conducts RL training on \textit{K\&K} to investigate how RL enhances reasoning capabilities.
\citet{he2024multiifbenchmarkingllmsmultiturn} provides evaluation through multilingual coverage and multi-turn design.
\citet{huang2025thinkbenchdynamicoutofdistributionevaluation, yu2025benchmarkingreasoningrobustnesslarge} test mathematical reasoning robustness through minimal perturbations.
Research by \citet{hazra2025largelanguagemodelslearned} investigates LLM reasoning capabilities through 3-SAT phase transitions.
Our work, \ours, uses randomly generated CNF instances to prevent data leakage and ensure continuity.
We provide five interrelated SAT-based \textit{problem types} and four \textit{question formats} with different information densities.
These three orthogonal dimensions—\textit{instance}, \textit{problem type}, and \textit{question format}—enable flexible experimental control for future LLM reasoning research.

\section{Conclusion}
\label{sec:conclusion}

We introduce \ours, a CNF-native verifier that systematically structures SAT-based reasoning evaluation and training along three orthogonal axes: \emph{instance}, \emph{problem type}, and \emph{question format}.
Our results reveal that while targeted reinforcement fine-tuning with verifiable rewards enables measurable gains—especially for harder or larger instances—persistent weaknesses remain in diagnostic reasoning and robustness to presentation changes.
The key patterns of generalization and their asymmetries are summarized in \cref{tab:generalization-findings}.

\textbf{Takeaway.} \ours provides a controllable and verifiable feedback loop for probing LLM logical reasoning.
It reliably exposes non-obvious failure modes—particularly format sensitivity and difficulty with minimal diagnoses—while enabling principled improvements without manual heuristics.
Future progress will require richer reward signals and explicit curriculum bridging across formats and domains, to close the gap between instance-specific improvements and generalizable, format-robust reasoning, and we release datasets and code to catalyze progress.


\section*{Limitations}

\paragraph{Scope and intent.}
\ours is designed as a \emph{controllable, verifiable, and factorized} probe of LLM logical reasoning along three axes (\textit{instance} / \textit{problem type} / \textit{question format}), prioritizing clean manipulation of variables over breadth.
\Pmcs/\Pmus are general diagnostic notions beyond propositional logic, but in this work they are \emph{instantiated} as clause-level subsets over CNF; richer logics (e.g., first-order, modal), multi-tool interaction, and open-ended natural-language tasks are out of scope.
The evaluation set fixes $m{=}4n$ with $n{\le}16$, the RFT training set uses $n{\le}8$, and narratives are synthetic and English-only.
Hence our conclusions target \emph{relative} patterns across instance/problem/format rather than absolute real-world capability.

\paragraph{Technical.}
By design, our setup prioritizes \emph{verifiable, controlled analysis}: models must return binary strings that we extract and check against the CNF with PySAT, which removes grading ambiguity but constrains the response space and can be brittle for long clause-mask outputs. 
All evaluations are conducted in a \emph{strict zero-shot} regime using standardized prompts (no tools, no few-shot exemplars), to isolate reasoning from prompt engineering; the prompt templates used for evaluation are provided in the paper and appendix.  
On the training side, we only instantiate reinforcement fine-tuning with the GRPO algorithm, optimizing a \emph{binary} (0/1) verifier reward; alternative policy-optimization methods, decoding strategies, and shaped rewards are not explored here.  
Task difficulty is proxied by classical solver statistics—decisions, conflicts, and propagations—which we compute with established solvers per task; the complete per-instance statistics and the solver identifiers are provided with the dataset metadata.

\paragraph{Narrative format robustness.}
Our conversions from CNF into the \Qstory and \Qdualstory formats use a fixed narrative template with entity names sampled from small predefined pools; we have not conducted systematic prompt-paraphrasing experiments to quantify how much the specific wording affects performance.
The consistent ordering \Qmath${}\ge{}$\Qdimacs${}>{}$\Qstory${}\approx{}$\Qdualstory across multiple models and tasks suggests that the main trend is not driven solely by a particular choice of phrasing, but we cannot rule out that alternative narrative framings would shift absolute accuracy.
More diverse narrative templates and controlled paraphrasing ablations are an important direction for future work.

\paragraph{Future work.}
Future directions include expanding the diversity and realism of question formats—incorporating both LLM-generated and programmatically varied representations such as tabular, code-based, or diagrammatic/logical schemas—to better capture the heterogeneity of human reasoning contexts.
We also plan to curate \emph{industrial-scale CNF corpora} drawn from real verification, planning, and scheduling benchmarks with standardized metadata for open research use.
On the learning side, designing \emph{dense and shaped rewards} (e.g., clause coverage, unit-propagation progress, \Pmcs/\Pmus cardinality margins, or partial-assignment consistency) may provide more informative optimization signals than binary correctness alone.
Curriculum learning across scale ($n,m$), problem-type mixtures, and structural regimes can help bridge small synthetic tasks and large real-world instances.
Finally, \ours is intended to serve as a practical foundation for \emph{empirical research} on LLM logical reasoning—offering a controllable, verifier-backed testbed for standardized and reproducible studies of reward shaping, curricula, and cross-format transfer.

\section*{Acknowledgments}
This research was partly supported by the Sichuan Science and Technology Program (2025JDDQ0008, 2024NSFJQ0035) and the Talents Program of the Sichuan Provincial Party Committee Organization Department.

\bibliography{satquest}

\appendix
\onecolumn

\section{Generate CNF Pair}
\label{app:generate_cnf_pair}

\begin{appendixtext}
The CNF generation algorithm is adapted from \citet{selsam2018learning}\footnote{\url{https://github.com/dselsam/neurosat/blob/master/python/gen_sr_dimacs.py}}.
The procedure takes four parameters: the number of variables~$n$, the number of clauses~$m$, the probability~$p_{k_2}$ of generating a unit clause, and the geometric distribution parameter~$p_{geo}$ that controls clause length.
It first builds an unsatisfiable CNF formula by repeatedly sampling random clauses until the conjunction is unsatisfiable (verified by a SAT solver).
It then derives a satisfiable partner by iteratively flipping literal polarities until the formula becomes satisfiable.
This paired generation ensures that each satisfiable instance has a closely related unsatisfiable counterpart, enabling controlled evaluation across both cases.
\end{appendixtext}

\begin{algorithm}
    \caption{Generate CNF Pair (Satisfiable and Unsatisfiable)}
    \label{alg:gen-cnf}
    \begin{algorithmic}[1]
    \Procedure{GenCNFPair}{$n, m, p_{k_2}, p_{geo}$}
    \State $unsat\_clauses \gets \emptyset$
    \While{$solve(unsat\_clauses)$} \Comment{Generate unsatisfiable CNF}
        \State $unsat\_clauses \gets \emptyset$
        \While{$|unsat\_clauses| < m$}
            \State $k \gets 1$ if $rand() < p_{k_2}$ else $2 + Geometric(p_{geo})$
            \State $k \gets \min(k, n)$
            \State $clause \gets RandomClause(n, k)$ \Comment{Random literals with polarity}
            \State $unsat\_clauses \gets unsat\_clauses \cup \{clause\}$
        \EndWhile
    \EndWhile
    \State $sat\_clauses \gets unsat\_clauses$
    \While{$\neg solve(sat\_clauses)$} \Comment{Convert to satisfiable CNF}
        \State $sat\_clauses \gets RandomFlipClause(sat\_clauses)$ \Comment{Flip literal polarities}
    \EndWhile
    \State \Return $unsat\_clauses, sat\_clauses$
    \EndProcedure
    \end{algorithmic}
\end{algorithm}

\newpage
\section{CNF to Narrative Conversion}
\label{app:cnf-to-narrative}

\begin{appendixtext}
This appendix details the deterministic procedure that converts a CNF instance into the \Qstory and \Qdualstory question formats.
The conversion preserves the logical structure of the original CNF while wrapping it in a cookie-day narrative.
A CNF object has two fields: \texttt{num\_vars}~($n$) and \texttt{clauses} (a list of clauses, each a list of integer literals in DIMACS convention, e.g.\ \texttt{[1, -2, 3]} for $x_1 \lor \lnot x_2 \lor x_3$).

\paragraph{Variable and entity mapping.}
Given $n$ variables and $m$ clauses, the procedure deterministically assigns: a chef name, $n$ cookie names (drawn from a predefined pool), and $m$ friend names.
Each variable $x_i$ maps to a cookie; positive literal $+i$ maps to the ``crunchy'' variant, and negative literal $-i$ maps to the ``chewy'' variant.

\paragraph{Story format (OR semantics).}
Each clause becomes a friend's preference list connected by disjunction: ``\textit{Friend is happy if she gets crunchy~A, chewy~B, or crunchy~C}.''
The full formula is satisfiable iff every friend can be made happy.

\paragraph{DualStory format (AND semantics).}
Each clause is negated and presented as a dislike combination connected by conjunction: ``\textit{Friend is unhappy only if she is served chewy~A and crunchy~B and chewy~C}.''
This requires the model to track De~Morgan's transformation.

\paragraph{Pseudocode.}
\cref{alg:cnf-to-story} gives the \Qstory conversion; the \Qdualstory variant differs only in the friend description template (replacing ``happy if \ldots\ or'' with ``unhappy only if \ldots\ and'') and swapping literal polarities in the phrase mapping.
The full object-oriented implementation covering all four formats and all five problem types is available in \texttt{satquest/question.py}.
\end{appendixtext}

\begin{algorithm}
\caption{CNF $\to$ \Qstory Prompt (simplified)}
\label{alg:cnf-to-story}
\begin{algorithmic}[1]
\Procedure{CnfToStory}{$\mathit{cnf}, \mathit{problem\_type}$}
\State $n, m \gets \mathit{cnf}.\mathit{num\_vars},\; |\mathit{cnf}.\mathit{clauses}|$
\State $\mathit{chef}, \mathit{cookies}[1..n], \mathit{friends}[1..m] \gets \Call{AssignNames}{n, m}$
\For{$i \gets 1$ \textbf{to} $n$} \Comment{Literal $\to$ phrase}
    \State $\mathit{phrase}[{+}i] \gets$ ``crunchy $\mathit{cookies}[i]$''
    \State $\mathit{phrase}[{-}i] \gets$ ``chewy $\mathit{cookies}[i]$''
\EndFor
\State $\mathit{lines} \gets$ [``It's cookie day! Chef \textit{chef} bakes $n$ cookies \ldots'']
\For{$k \gets 1$ \textbf{to} $m$} \Comment{Clause $\to$ friend preference}
    \State $\mathit{opts} \gets [\mathit{phrase}[\ell]$ for $\ell \in \mathit{cnf}.\mathit{clauses}[k]]$
    \State Append ``$k.$ $\mathit{friends}[k]$ wants: $\mathit{opts}[1],\ldots$'' to $\mathit{lines}$
\EndFor
\State Append task instruction for $\mathit{problem\_type}$ to $\mathit{lines}$
\State \Return $\mathit{lines}$ joined by newline
\EndProcedure
\end{algorithmic}
\end{algorithm}

\newpage
\section{Prompt Details}
\label{app:prompt-details}

\begin{figure}[ht]
    \centering
    \includegraphics[width=1.0\linewidth]{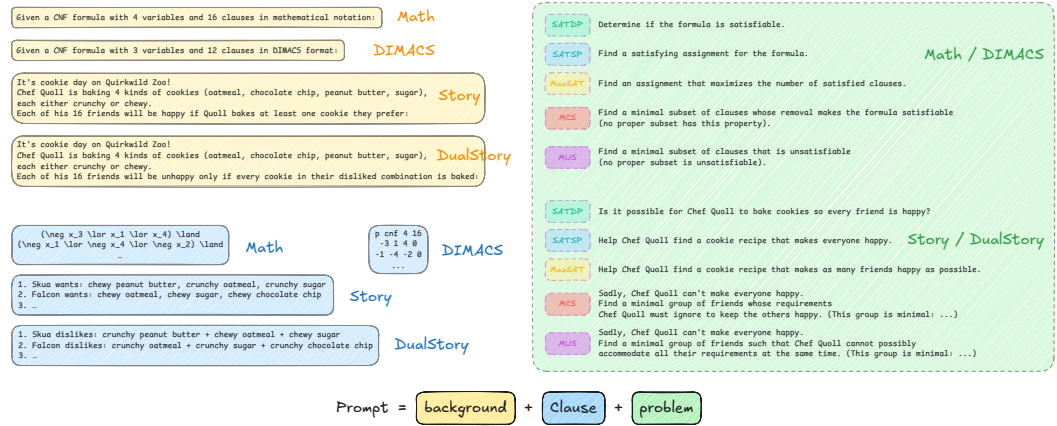}
    \caption{Task Prompt Construction in \ours}\label{fig:prompt}
\end{figure}

\begin{appendixtext}
In this appendix, we present the complete prompt templates and representative example outputs used to instruct LLMs to produce binary string responses for SAT-based reasoning tasks.
These templates specify the precise wording, formatting requirements, and task descriptions used during both evaluation and reinforcement fine-tuning phases.
Our prompt construction strategy adapts and extends established templates from OpenAI's \texttt{simple-eval}\footnote{\url{https://github.com/openai/simple-evals/blob/0f2cf3/math_eval.py}} and \texttt{DeepSeek-R1}~\citep{deepseekai2025deepseekr1incentivizingreasoningcapability}.
The following subsections detail how we systematically assemble task prompts and incorporate system-level configurations for both evaluation (\cref{sec:evaluations}) and fine-tuning (\cref{sec:rft}) experiments.
\cref{fig:prompt} illustrates the general structure of the task prompt used throughout our experiments.
\end{appendixtext}

\subsection{Evaluation Prompt Template}

In \cref{sec:evaluations}, we employ the following prompt template that includes Chain-of-Thought (CoT) guidance and specific output format instructions to ensure consistent responses across all models.

For vanilla model, add system prompt: \texttt{"You are a helpful assistant."}.

\begin{tcolorbox}[llmstyle, label=lst:eval-prompt-template]
\begin{lstlisting}
Solve the following problem step by step. The last line of your response should be of the form Answer: $ANSWER (without quotes) where $ANSWER is the answer to the problem.

Given a CNF formula with 4 variables and 16 clauses in DIMACS format:

p cnf 4 16
-3 1 4 0
-1 -4 -2 0
-1 4 2 0
2 -1 -4 0
2 -3 4 0
-3 -4 1 0
-4 1 -3 0
1 2 -4 0
-3 -2 1 0
4 -3 1 0
-1 -3 2 0
2 -3 4 0
-1 -2 3 0
2 3 4 0
2 3 1 0
1 3 -4 0

Find a satisfying assignment for the formula.
Output a binary string of length 4 ('1' for true, '0' for false).

Remember to put your answer on its own line after "Answer:", and you do not need to use a \boxed command.
\end{lstlisting}
\end{tcolorbox}

\newpage
\subsection{Reinforcement Fine-Tuning Prompt Template}

For \cref{sec:rft}, we adapt the prompt to encourage explicit CoT reasoning by combining system instructions with a structured user message format.

\begin{tcolorbox}[llmstyle, label=lst:rft-prompt-template]
\begin{lstlisting}
<|im_start|>system
You are a helpful AI Assistant that provides well-reasoned and detailed responses. You first think about the reasoning process as an internal monologue and then provide the user with the answer. Respond in the following format: <think>\n...\n</think>\n<answer>\n...\n</answer>
<|im_end|>
<|im_start|>user
Given a CNF formula with 4 variables and 16 clauses in DIMACS format:

p cnf 4 16
-3 1 4 0
-1 -4 -2 0
-1 4 2 0
2 -1 -4 0
2 -3 4 0
-3 -4 1 0
-4 1 -3 0
1 2 -4 0
-3 -2 1 0
4 -3 1 0
-1 -3 2 0
2 -3 4 0
-1 -2 3 0
2 3 4 0
2 3 1 0
1 3 -4 0

Find a satisfying assignment for the formula.
Output a binary string of length 4 ('1' for true, '0' for false).

Show your work in <think> </think> tags. And return the final answer in <answer> </answer> tags, for example <answer> 0101 </answer>.
<|im_end|>
<|im_start|>assistant
\end{lstlisting}
\end{tcolorbox}

\clearpage

\section{Prompt Examples}
\label{app:prompt-examples}

For more prompt examples, please refer to the \url{https://satquest.sdpkjc.com/} website.

\begin{tcolorbox}[llmstyle, title=\Psatsp-\Qmath Prompt Example, label=lst:prompt-example-math]
\begin{lstlisting}
Given a CNF formula with 3 variables and 4 clauses in mathematical notation:

(x_1 \lor \neg x_2 \lor x_3) \land (\neg x_1 \lor x_2 \lor x_3) \land (x_1 \lor x_2) \land (\neg x_3)

Find a satisfying assignment for the formula.
Output a binary string of length 3 ('1' for true, '0' for false).
\end{lstlisting}
\end{tcolorbox}

\begin{tcolorbox}[llmstyle, title=\Psatsp-\Qdimacs Prompt Example, label=lst:prompt-example-dimacs]
\begin{lstlisting}
Given a CNF formula with 3 variables and 4 clauses in DIMACS format:

p cnf 3 4
1 -2 3 0
-1 2 3 0
1 2 0
-3 0

Find a satisfying assignment for the formula.
Output a binary string of length 3 ('1' for true, '0' for false).
\end{lstlisting}
\end{tcolorbox}

\begin{tcolorbox}[llmstyle, title=\Psatsp-\Qstory Prompt Example, label=lst:prompt-example-story]
\begin{lstlisting}
It's cookie day on Quirkwild Zoo!
Chef Orion is baking 3 kinds of cookies (chocolate comet, maple moon, ginger nebula), each either crunchy or chewy.
Each of his 4 friends will be happy if Orion bakes at least one cookie they prefer:


1. Aquila wants: crunchy chocolate comet, chewy maple moon, crunchy ginger nebula
2. Borealis wants: chewy chocolate comet, crunchy maple moon, crunchy ginger nebula
3. Cygnus wants: crunchy chocolate comet, crunchy maple moon
4. Draco wants: chewy ginger nebula

Find a satisfying assignment for the formula.
Output a binary string of length 3 ('1' for true, '0' for false).
\end{lstlisting}
\end{tcolorbox}

\begin{tcolorbox}[llmstyle, title=\Psatsp-\Qdualstory Prompt Example, label=lst:prompt-example-dualstory]
\begin{lstlisting}
It's cookie day on Quirkwild Zoo!
Chef Orion is baking 3 kinds of cookies (chocolate comet, maple moon, ginger nebula), each either crunchy or chewy.
Each of his 4 friends will be unhappy only if every cookie in their disliked combination is baked:


1. Aquila dislikes: chewy chocolate comet + crunchy maple moon + chewy ginger nebula
2. Borealis dislikes: crunchy chocolate comet + chewy maple moon + chewy ginger nebula
3. Cygnus dislikes: chewy chocolate comet + chewy maple moon
4. Draco dislikes: crunchy ginger nebula

Find a satisfying assignment for the formula.
Output a binary string of length 3 ('1' for true, '0' for false).
\end{lstlisting}
\end{tcolorbox}

\clearpage

\section{Case Studies}
\label{app:case-studies}

\begin{appendixtext}
Our analysis reveals that both \texttt{DeepSeek-R1} and \texttt{QwQ-32B} predominantly reason directly with the \Qdimacs format rather than translating it into formal mathematical notation.
Throughout the four case studies, segments highlighted in green demonstrate instances where the models correctly comprehend the CNF structure, especially regarding the disjunctive (\texttt{OR}) semantics within clauses.
Conversely, red-highlighted sections expose critical reasoning flaws: models frequently misinterpret the disjunctive nature of clauses or conflate the semantic meaning of positive and negative literals (representing true versus false values).
These observations suggest that direct reasoning within the \Qdimacs representation, without proper symbolic abstraction, frequently results in semantic drift and format-specific reasoning errors.
\end{appendixtext}

\begin{tcolorbox}[llmstyle, title=Case1: DeepSeek-R1 @ \Psatsp-\Qdimacs, label=lst:case-study-example1]
\begin{lstlisting}
Okay, so I need to find a satisfying assignment for this CNF formula with 9 variables and 36 clauses.

...

(*@\colorbox{highlightgreen}{So, for each clause, at least one of the literals must be true in the assignment.}@*)

...

For each variable, check if all occurrences are positive, all negative, or mixed.

Let's start with variable 1:

Looking through the clauses where variable 1 appears.

Clause 5: -1 2 -6 5 8. So -1 here.

Clause 6:8 1 9 -6. Positive 1.

...

(*@\colorbox{highlightred}{Clause1: -2 -9 6 -> with 6=0, this becomes -2 AND -9 -> so both must be true. So variables2 and 9 must be 0.}@*)

...

All clauses seem to be satisfied. So the assignment is valid.    
\end{lstlisting}
    \tcblower 
    \small Full prompt and response: \blindurl{https://gist.github.com/sdpkjc/3e3df5e9f268588dab862e795b080258}
\end{tcolorbox}

\begin{tcolorbox}[llmstyle, title=Case2: DeepSeek-R1 @ \Psatsp-\Qdimacs, label=lst:case-study-example2]
\begin{lstlisting}
Okay, so I need to find a satisfying assignment for this CNF formula with 10 variables and 40 clauses. Hmm, let's think. The variables are numbered 1 through 10, and each line in the DIMACS format represents a clause.
(*@\colorbox{highlightgreen}{The clauses are ORs of literals, and the entire formula is the AND of all these clauses.}@*)
(*@\colorbox{highlightgreen}{So the assignment has to make at least one literal in each clause true.}@*)

...

Clause 5: -1 -> negative occurrence.

Clause 6: 1 -> positive.

...

Ok, moving on.

(*@\colorbox{highlightred}{Clause28:-5 -1 -> both must be false. So variable5=0 and variable1=0.}@*)

Wait, ...

All other clauses seem to have been handled earlier. So this assignment should work. Thus, the binary string is 0001010100.
       
\end{lstlisting}
\tcblower 
\small Full prompt and response: \blindurl{https://gist.github.com/sdpkjc/ab0963d08dc65cb9352e4fa489b39040}
\end{tcolorbox}

\begin{tcolorbox}[llmstyle, title=Case3: QwQ-32B @ \Psatsp-\Qdimacs, label=lst:case-study-example3]
\begin{lstlisting}
Okay, so I have this CNF formula with 9 variables and 36 clauses, and I need to find a satisfying assignment. 

...

Let me start by writing down some of the clauses:

(*@\colorbox{highlightred}{1. 5 3 -4 9 0 -> variables 5,3 are true, 4 is false, 9 is true.}@*)
(*@\colorbox{highlightred}{2. -1 -2 7 -6 0 -> variables 1 and 2 are false, 7 and 6 are false?}@*)
(*@\colorbox{highlightgreen}{Wait, no, the clause is satisfied if at least one literal is true.}@*)
So for this clause, either -1 (1 is false), -2 (2 false), 7 (true), or -6 (6 false). So ...

...

So this should be a valid assignment.
\end{lstlisting}
\tcblower 
\small Full prompt and response: \blindurl{https://gist.github.com/sdpkjc/415a9f0d6e9166f9ca6c19f4bbb5be4f}
\end{tcolorbox}

\begin{tcolorbox}[llmstyle, title=Case4: QwQ-32B @ \Psatsp-\Qdimacs, label=lst:case-study-example4]
\begin{lstlisting}
Okay, so I need to find a satisfying assignment for this CNF formula with 10 variables and 40 clauses. Hmm, that's a bit of a challenge.
Let me think about how to approach this.
(*@\colorbox{highlightgreen}{Since it's a CNF formula, each clause has to be satisfied by at least one literal being true.}@*)

...

Looking at the clauses:

The first few clauses are:

-9 4 -10 0 -> clause 1

6 -2 -8 -1 0 -> clause 2

....

(*@\colorbox{highlightred}{Clause4: -1 -2 -4 -> since variables1 and2 are 0, this clause requires -4 (so variable4 must be 0).}@*)

...

Yes, that works.

Thus, the satisfying assignment is 1000101101.
\end{lstlisting}
\tcblower 
\small Full prompt and response: \blindurl{https://gist.github.com/sdpkjc/2aef69a7b663be4e0ca5325b7526ab5d}
\end{tcolorbox}

\newpage

\section{Evaluation Configs}
\label{app:evaluation-configs}

In all experiments in \cref{sec:evaluations}, we use APIs for evaluation. The specific model APIs and IDs are listed in \cref{tab:model-api-id}, and the parameters for vanilla models are shown in \cref{tab:evaluation-params}.
For reasoning models, we use their default configurations.

\begin{table}[ht]
    \centering
    \caption{Model API / ID of our Evaluated Models.}
    \label{tab:model-api-id}
    \begin{tabular}{l l}
        \toprule
        \textbf{Model Name} & \textbf{API / ID} \\
        \midrule
        \oThreeMini & Azure API\tablefootnote{\textit{Azure OpenAI API}: \url{https://azure.microsoft.com/en-us/products/ai-services/openai-service}}: \texttt{o3-mini-2025-01-31} \\
        \GPT & Azure API: \texttt{gpt-4.1-2025-04-14} \\
        \DSR & VolcEngine API\tablefootnote{\textit{ByteDance VolcEngine AI platform API}: \url{https://www.volcengine.com/}}: \texttt{deepseek-r1} \\
        \DSV & VolcEngine API: \texttt{deepseek-v3-0324} \\
        \DSRDistillQwenSmall & VolcEngine API: \texttt{deepseek-r1-distill-qwen-7b} \\
        \DSRDistillQwenLarge & VolcEngine API: \texttt{deepseek-r1-distill-qwen-32b} \\
        \QwQ & Alibaba Cloud API: \texttt{qwq-32b-plus}\tablefootnote{\textit{Alibaba Cloud API}: \url{https://www.alibabacloud.com/en}} \\
        \QwenSmall & Alibaba Cloud API: \texttt{qwen2.5-7b-instruct} \\
        \QwenLarge & Alibaba Cloud API: \texttt{qwen2.5-32b-instruct} \\
        \bottomrule
    \end{tabular}
\end{table}

\begin{table}[ht]
    \centering
    \caption{Evaluation Parameters for Vanilla Models}
    \label{tab:evaluation-params}
    \begin{tabular}{l l}
        \toprule
        \textbf{Parameter} & \textbf{Value} \\
        \midrule
        \texttt{temperature} & \(0.6\) \\
        \texttt{top\_p} & \(1.0\) \\
        \texttt{max\_tokens} & \(16384\) \\
        \bottomrule
    \end{tabular}
\end{table}

\clearpage
\newpage

\section{Reinforcement Fine-Tuning Configs}
\label{app:rft-configs}

\subsection{Format Reward Functions}
\label{app:format-reward}

The format reward functions are adapted from the \texttt{huggingface/Open-R1} library\footnote{\url{https://github.com/huggingface/open-r1/blob/main/src/open_r1/rewards.py}}.

\begin{tcolorbox}[llmstyle, label=lst:format-reward]
\begin{lstlisting}
def tag_count_reward(completions, **kwargs) -> list[float]:
    def count_tags(text: str) -> float:
        count = 0.0
        if text.count("<think>") == 1:
            count += 0.25
        if text.count("</think>") == 1:
            count += 0.25
        if text.count("<answer>") == 1:
            count += 0.25
        if text.count("</answer>") == 1:
            count += 0.25
        return count

    contents = [completion[0]["content"] for completion in completions]
    return [count_tags(c) for c in contents]

def format_reward(completions, **kwargs):
    _PATTERN = re.compile(r"<think>.*?</think>\s?<answer>.*?</answer>", flags=re.DOTALL)
    completion_contents = [completion[0]["content"] for completion in completions]
    rewards = []
    for c in completion_contents:
        text = str(c)
        total_len = len(text)
        if total_len == 0:
            rewards.append(0.0)
            continue

        m = _PATTERN.search(text)
        match_len = len(m.group()) if m else 0
        rewards.append(match_len / total_len)

    return rewards
\end{lstlisting}
\end{tcolorbox}

\subsection{Training Parameters}
\label{app:training-params}
The training parameters used for GRPO of the \texttt{Qwen2.5-7B-Instruct} model are summarized in \cref{tab:training-params}.

\begin{table}[ht]
    \centering
    \caption{Training Parameters for GRPO}
    \label{tab:training-params}
    \begin{tabular}{l l}
        \toprule
        \textbf{Parameter} & \textbf{Value} \\
        \midrule
        \texttt{learning\_rate} & \(0.000002\) \\
        \texttt{batch\_size} & \(\texttt{num\_generations} \times 8 = 128\) \\
        \texttt{max\_grad\_norm} & \(0.3\) \\
        \texttt{num\_iterations} & \(1\) \\
        \texttt{beta} & \(0.01\) \\
        \texttt{max\_steps} & \(500\) \\
        \texttt{max\_prompt\_length} & \(2048\) \\
        \texttt{max\_completion\_length} & \(8192\) \\
        \texttt{mask\_truncated\_completions} & \texttt{True} \\
        \texttt{num\_generations} & \(16\) \\
        \texttt{temperature} & \(1.0\) \\
        \texttt{scale\_rewards} & \texttt{True} \\
        \bottomrule
    \end{tabular}
\end{table}

\subsection{Experiments Compute Resources}
\label{app:exp-compute-resources}

All experiments were conducted on a single server node equipped with \(8\) \texttt{NVIDIA A100 80GB} GPUs, \(2\) \texttt{Intel Xeon Platinum 8350C} CPU, and \texttt{1600GB} memory.
We allocated \(4\) GPUs for training and \(4\) GPUs for \textit{VLLM} inference.
The training time for \texttt{GRPO@\Psatsp-\Qmath}, \texttt{GRPO@\Pmaxsat-\Qmath}, and \texttt{GRPO@\Psatsp-\Qstory} was approximately \(30\) hours, \(26\) hours, and \(9\) hours, respectively.

\clearpage
\newpage

\section{Additional Figures}
\label{app:additional-figures}

\begin{figure}[ht]
    \centering
    \includegraphics[width=0.9\linewidth]{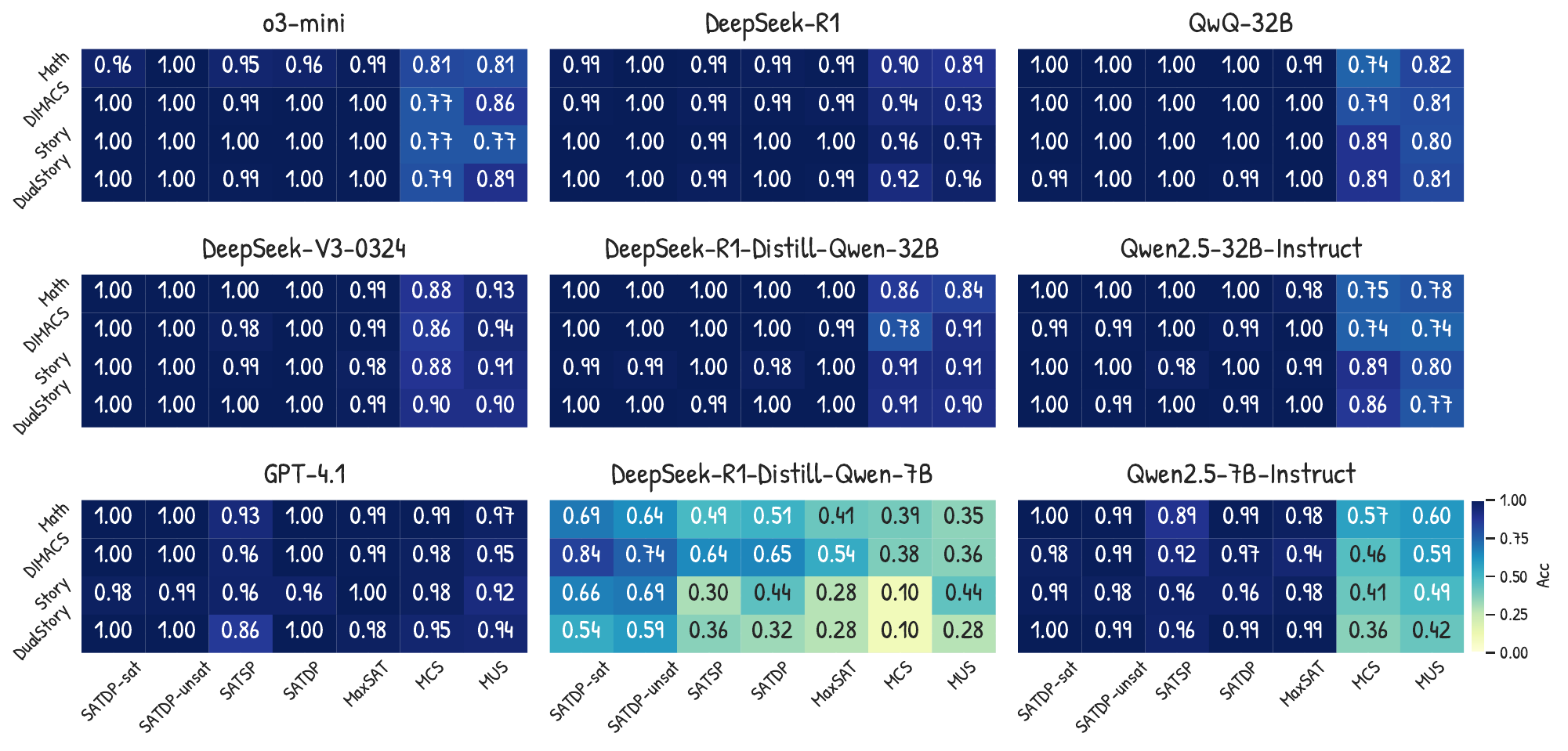}
    \caption{Format accuracy heatmaps showing LLM performance breakdown by \textit{problem type} (columns) and \textit{question format} (rows).}\label{fig:heatmaps_format}
\end{figure}

\begin{figure}[ht]
    \centering
    \includegraphics[width=0.9\linewidth]{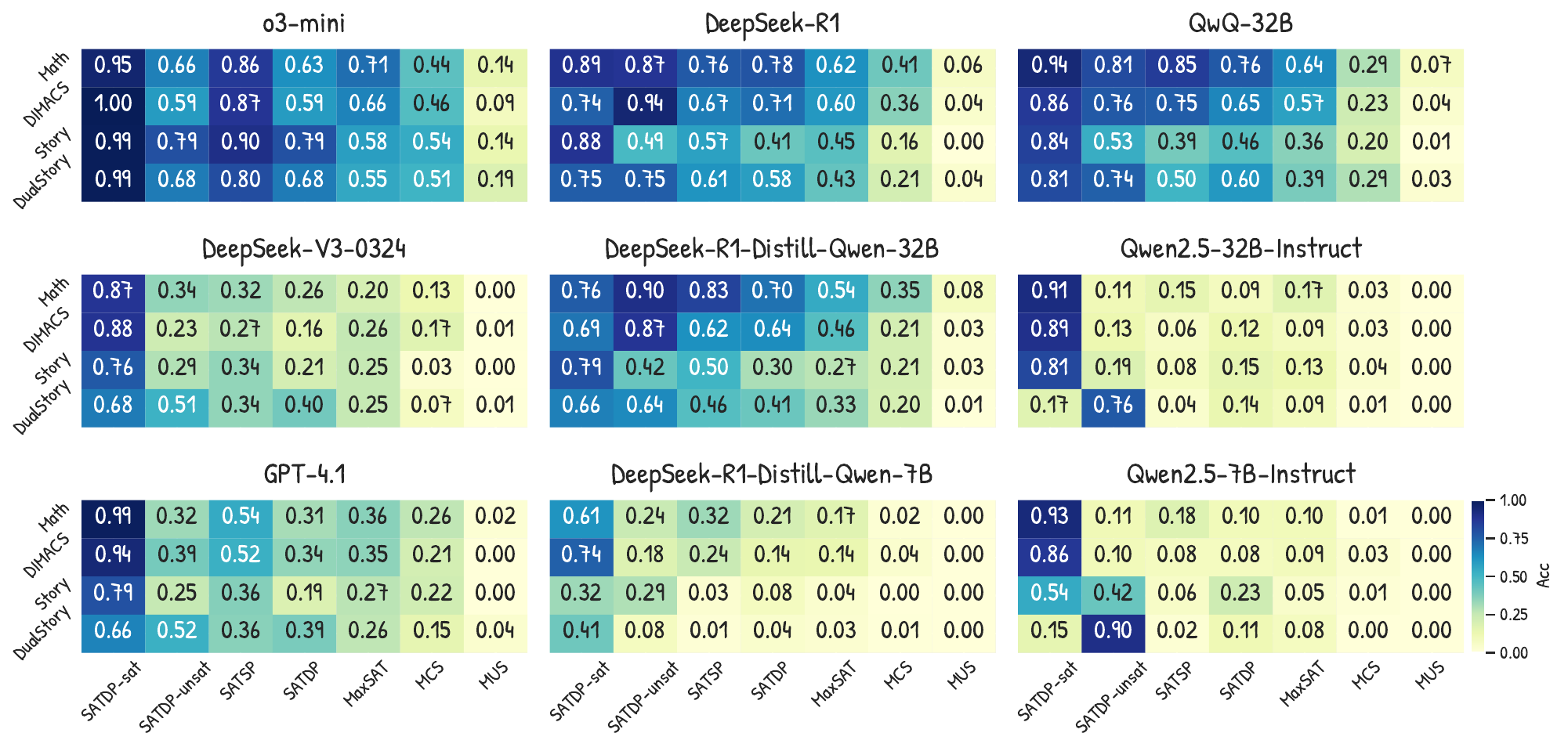}
    \caption{All accuracy heatmaps (including \Psatdpsat and \Psatdpunsat) showing LLM performance breakdown by \textit{problem type} (columns) and \textit{question format} (rows).}\label{fig:heatmaps-all}
\end{figure}

\clearpage

\section{Hallucination Audit}
\label{app:hallucination-audit}

\begin{appendixtext}
To substantiate the claim that high-complexity instances trigger hallucination-dominated failures across models, we conducted a dedicated error-type audit following the protocol below.

\paragraph{Protocol.}
From the \ourDataset evaluation set (140 CNF pairs), we selected the 20 pairs whose pair ID is divisible by~7.
For each pair we instantiated four \Qmath-format tasks (\Psatsp, \Pmaxsat, \Pmcs, \Pmus), yielding 80 instances.
We collected responses from three reasoning models: \oThreeMini, \QwQ, and \DSR.
Error types were labelled by \texttt{Claude-Sonnet-4.5} in an LLM-as-a-Judge protocol, with 10\% of cases manually audited by a human expert to verify label quality.

\paragraph{Error taxonomy.}
We adopt a multi-label taxonomy (a single incorrect response may exhibit multiple types):
\begin{itemize}[leftmargin=1em]
  \item \textbf{Type~A} (Logic Error): a mostly honest, step-by-step reasoning attempt with a concrete calculation or tracking mistake.
  \item \textbf{Type~B} (Loop): repetitive reasoning without real progress.
  \item \textbf{Type~C} (Fabricated Execution): the model fabricates intermediate solver states or verification steps (e.g., asserting that a clause is satisfied when it is not).
  \item \textbf{Type~D} (Invented Shortcut): the model invents non-existent logical rules or ``shortcuts'' to bypass the combinatorial reasoning required.
\end{itemize}

\paragraph{Results by task.}
\cref{tab:hallucination-by-task} summarizes the failure-mode distribution.
Across all three models, complex diagnostic/optimization tasks (\Pmaxsat, \Pmcs, \Pmus) are dominated by Type~C hallucination (fabricated execution), often supplemented by Type~D shortcuts.
Simple logic mistakes without fabrication (Type~A) account for only a small fraction of errors on these tasks.
\end{appendixtext}

\begin{table}[h]
\centering
\small
\caption{Failure-mode distribution by task (\% of total samples). Categories are non-exclusive (multi-label); rows may sum to ${>}100\%$.}
\label{tab:hallucination-by-task}
\begin{tabular}{ll rrrr r}
\toprule
\textbf{Model} & \textbf{Task} & \textbf{A} & \textbf{B} & \textbf{C} & \textbf{D} & \textbf{C$\cup$D} \\
\midrule
\oThreeMini & \Psatsp  &  5.0 & 0.0 & 10.0 &  0.0 & 10.0 \\
            & \Pmaxsat & 45.0 & 0.0 & 45.0 &  0.0 & 45.0 \\
            & \Pmcs    &  5.0 & 0.0 & 75.0 & 10.0 & 75.0 \\
            & \Pmus    &  0.0 & 0.0 & 50.0 &  0.0 & 50.0 \\
\midrule
\QwQ        & \Psatsp  & 35.0 & 0.0 & 60.0 &  0.0 & 60.0 \\
            & \Pmaxsat & 15.0 & 0.0 & 25.0 &  0.0 & 25.0 \\
            & \Pmcs    &  0.0 & 0.0 & 95.0 &  5.0 &100.0 \\
            & \Pmus    &  0.0 & 0.0 &100.0 &  0.0 &100.0 \\
\midrule
\DSR        & \Psatsp  & 20.0 & 0.0 & 65.0 &  0.0 & 65.0 \\
            & \Pmaxsat & 10.0 & 0.0 & 75.0 &  0.0 & 75.0 \\
            & \Pmcs    &  5.0 & 0.0 & 95.0 & 40.0 & 95.0 \\
            & \Pmus    & 35.0 & 0.0 & 70.0 & 30.0 & 80.0 \\
\bottomrule
\end{tabular}
\end{table}

\begin{appendixtext}
\paragraph{Hallucination vs.\ complexity.}
\cref{tab:hallucination-by-complexity} buckets instances by solver decisions and reports the hallucination rate (Type~C or~D) within each bucket.
The pattern is consistent across architectures: as solver-measured complexity increases, the fraction of failures attributed to hallucination rises sharply.
\end{appendixtext}

\begin{table}[h]
\centering
\small
\caption{Hallucination rate (Type~C$\cup$D, \% of total samples) vs.\ instance complexity (solver decisions).}
\label{tab:hallucination-by-complexity}
\begin{tabular}{l rrr}
\toprule
\textbf{Complexity (decisions)} & \textbf{\oThreeMini} & \textbf{\QwQ} & \textbf{\DSR} \\
\midrule
Low (3--10)    & 10.3 & 51.7 & 65.5 \\
Medium (10--76)  & 60.0 & 64.0 & 84.0 \\
High (76--903) & 69.2 &100.0 & 88.5 \\
\bottomrule
\end{tabular}
\end{table}

\begin{appendixtext}
These results provide quantitative evidence that, under low complexity, models sometimes fail due to ordinary logic errors (Type~A), whereas under medium and especially high complexity, failures shift to hallucinated solver states or invented reasoning shortcuts (Types~C/D).
This shift appears consistently for all three models, even though their overall accuracy levels differ.
All sampling scripts, raw model outputs, and judge logs are released with the \ours repository.
\end{appendixtext}


\section{Additional Experiments}
\label{app:additional-experiments}

\begin{appendixtext}
\paragraph{Scaling to Larger Models.}
We applied the same RFT protocol to \texttt{Qwen2.5-14B-Instruct}, following the GRPO setup used for the 7B model. As shown in \cref{tab:additional14b}, the 14B model exhibits consistent improvements on all in-domain tasks, confirming \ours's scalability. However, the cross-format generalization gap persists, suggesting that the limitation stems from current RFT methods rather than model size.

\paragraph{Generalization Beyond SATQuest.}
To examine whether \ours training benefits broader reasoning tasks, we evaluated fine-tuned models on two external benchmarks: \textit{AIME2024} (pass@4) and \textit{DeepMind Mathematics}.
As summarized in \cref{tab:transfer}, \ours-trained models show consistent improvements across both datasets, indicating enhanced mathematical reasoning rather than overfitting to \ours formats.

\paragraph{Summary.}
These extended results confirm that: (1) \ours-derived RFT scales to larger models; (2) improvements extend to independent reasoning benchmarks; and (3) the remaining cross-format gap reflects an intrinsic challenge in current RFT methods rather than dataset bias or model capacity.
\end{appendixtext}

\begin{table}[h]
\centering
\small
\begin{tabular}{lccc}
\toprule
Model & Evaluation on \Psatsp-\Qmath & Evaluation on \Psatsp-\Qstory & Evaluation on \Pmaxsat-\Qmath \\
\midrule
\texttt{Qwen2.5-14B-Instruct} & 12.86 $\pm$ 1.91 & 4.38 $\pm$ 1.77 & 12.94 $\pm$ 1.35 \\
+ GRPO@\Psatsp-\Qmath & 70.80 $\pm$ 3.14 & 0.63 $\pm$ 0.80 & 17.05 $\pm$ 1.59 \\
+ GRPO@\Psatsp-\Qstory & 11.96 $\pm$ 2.83 & 11.25 $\pm$ 1.31 & 14.20 $\pm$ 3.18 \\
+ GRPO@\Pmaxsat-\Qmath & 41.52 $\pm$ 2.30 & 1.25 $\pm$ 0.63 & 46.33 $\pm$ 2.24 \\
+ GRPO@\Psatsp-\Qmath/\Qstory & 64.38 $\pm$ 3.71 & 10.54 $\pm$ 2.08 & 21.96 $\pm$ 2.88 \\
\bottomrule
\end{tabular}
\caption{Additional RFT results on \texttt{Qwen2.5-14B-Instruct} (mean $\pm$ std over 8 runs).}
\label{tab:additional14b}
\end{table}

\begin{table}[h]
\centering
\small
\begin{tabular}{lcc}
\toprule
Model & \textit{AIME2024} (pass@4) & \textit{DeepMind Mathematics} \\
\midrule
\texttt{Qwen2.5-14B-Instruct} & 24.17 $\pm$ 5.69 & 75.70 $\pm$ 0.74 \\
+ GRPO@\Psatsp-\Qmath & 25.83 $\pm$ 3.19 & 79.15 $\pm$ 0.95 \\
+ GRPO@\Psatsp-\Qstory & 20.84 $\pm$ 4.19 & 79.15 $\pm$ 0.55 \\
+ GRPO@\Pmaxsat-\Qmath & 24.16 $\pm$ 1.67 & 77.72 $\pm$ 0.49 \\
+ GRPO@\Psatsp-\Qmath/\Qstory & 25.00 $\pm$ 4.30 & 79.30 $\pm$ 0.55 \\
\bottomrule
\end{tabular}
\caption{Out-of-domain generalization on external reasoning benchmarks.}
\label{tab:transfer}
\end{table}

\end{document}